\documentclass{article}
\usepackage{iclr2027_conference,times}

\usepackage{amsmath,amsfonts,bm}

\def\eqref#1{equation~\ref{#1}}
\def\Eqref#1{Equation~\ref{#1}}

\def\1{\bm{1}}

\DeclareMathAlphabet{\mathsfit}{\encodingdefault}{\sfdefault}{m}{sl}
\SetMathAlphabet{\mathsfit}{bold}{\encodingdefault}{\sfdefault}{bx}{n}

\usepackage{hyperref}
\usepackage{url}
\usepackage{amsmath,amssymb}
\usepackage{bm}
\usepackage{graphicx}
\usepackage{booktabs}
\usepackage{multirow}
\usepackage{dsfont}
\usepackage{amsthm}
\usepackage{tikz}
\usepackage{pgfplots}
\pgfplotsset{compat=1.18}
\usetikzlibrary{arrows.meta,shapes.geometric}
\newtheorem{proposition}{Proposition}

\iclrfinalcopy

\title{Style-Debiased DPO:\\ Updating LLM Knowledge with\\ Factuality-Aware Synthetic Preference Data}

\author{Takayuki Yamamoto \quad Daisuke Kawahara\\
Waseda University\\
\texttt{takayukiyamamoto@ruri.waseda.jp}
}

\newcommand{\thr}{f_{\text{thr}}}
\newcommand{\fmax}{f_{\max}}
\newcommand{\piref}{\pi_{\text{ref}}}

\begin{document}
\maketitle
\lhead{Preprint}

\begin{abstract}
Continued pretraining (CPT) with data augmentation such as paraphrasing can store inside a large language model (LLM) the knowledge of a small source corpus, a collection of documents describing new knowledge. The stored knowledge, however, is not always retrieved correctly. We study the eliciting side rather than the storing side: we use preference optimization, which learns from pairs of a preferred (chosen) and a dispreferred (rejected) response, so that the model \emph{elicits} its stored knowledge more accurately. One proposed approach takes the model's own erroneous response as rejected and the gold answer as chosen, so as to suppress the error. When the target knowledge is \emph{partially known}, however, most of these rejected responses are factually correct. Using direct preference optimization (DPO) as the preference optimization method then pushes down rejected responses that contain correct knowledge and differ from the chosen answer only in style, such as length and wording. We propose \emph{style-debiased DPO} (SD-DPO), which scores whether the rejected response of each pair is factually correct, inverts the preference of such pairs, and weights them so that the learning signal due to differences in style cancels out as a whole. We first test whether, on top of EntiGraph, a representative storing-side method that runs CPT on text synthesized from the corpus, our method adds accuracy efficiently. On QuALITY, the reading-comprehension QA benchmark on which EntiGraph was evaluated, SD-DPO exceeds a baseline we CPT on EntiGraph's synthetic data from the same base model and evaluate with the same procedure. The training tokens this requires are a few dozen times fewer than the additional CPT needed for the same gain. For knowledge updating, the main goal of this work, we use AToKE, a knowledge-editing benchmark for facts that change over time. There, SD-DPO reaches an overall accuracy of 0.982 and answers with the new or the old fact according to the queried period.
\end{abstract}

\section{Introduction}
\label{sec:intro}

Teaching a pretrained large language model (LLM) new knowledge remains difficult in several respects. New facts are learned more slowly than facts the model already knows, and must be presented repeatedly in diverse paraphrases~\citep{allen2024physics,ovadia2024finetuning,chang2024how}. Training on a document does not guarantee that its content can be elicited as answers to questions~\citep{berglund2024reversal,jiang2024instruction}. Learning new knowledge increases hallucination~\citep{gekhman2024does}. The new knowledge also interferes with existing abilities and with old knowledge~\citep{luo2023empirical,xu2024knowledge}. Methods that let an LLM answer correctly about new knowledge fall into two groups. One keeps the documents describing the new knowledge outside the model and supplies them as context at inference time through retrieval-augmented generation (RAG;~\citealp{lewis2020retrieval}). The other performs \emph{continued pretraining} (CPT) on the collection of such documents (henceforth the \emph{source corpus}), updating the model parameters directly so that the LLM itself can answer. This paper concerns the latter. The naive form, CPT on the source corpus as is, has been reported to lower accuracy when the corpus is small, because the phrasings are not diverse and the distribution is narrow~\citep{yang2025entigraph}; data augmentation such as paraphrasing is therefore used. Its representative, EntiGraph~\citep{yang2025entigraph}, generates synthetic texts that explain the relations among the entities of the source corpus in many different phrasings, and showed that training on them makes closed-book QA accuracy (answering from memory, without the source text) scale log-linearly with the number of training tokens. Similarly, PASTA~\citep{yamamoto2026pasta} combines paraphrase-based augmentation of the source corpus, QA generation over the source corpus, and self-preference optimization into a method that integrates the knowledge of news articles past the LLM's knowledge cutoff into the LLM. However, the EntiGraph scaling curve also exposes a limit: as training proceeds, each additional token teaches less, and accuracy levels off. Storing more knowledge does not by itself make a task that asks about that knowledge \emph{elicit} correct answers in proportion. That is, data augmentation has made storing knowledge feasible, but the difficulty of eliciting the stored knowledge correctly, noted above, remains.

We therefore treat acquiring knowledge and eliciting it correctly as two separate problems, and adopt a two-stage design that places a preference-optimization stage after CPT. We assume that the main role of the first stage, CPT, is to \emph{store} knowledge, and that the main role of the second stage, preference optimization, is to make it \emph{retrievable}. Our experiments confirm this assumption (Section~\ref{sec:results_quality}). For preference optimization we use direct preference optimization (DPO;~\citealp{rafailov2023direct}), which pairs each question with a chosen and a rejected response, raises the likelihood of the former, and lowers that of the latter. The responses we want to suppress are the erroneous ones the model currently produces, and those are obtained simply by sampling the model itself; we therefore take the gold answer as chosen and the CPT model's own sampled response as rejected---the recipe of PASTA's self-preference optimization (Self-DPO) and of the method of \citet{rozner2024knowledge}, which adapts the DPO objective to knowledge editing.

Preference optimization, however, learns any feature that differs systematically between chosen and rejected responses, even one unrelated to content. Response length is the best-known example~\citep{park2024disentangling,lu2024sampo}, and more generally any feature whose mean differs between the two is learned from the first gradient step~\citep{moya2026spurious}. Existing remedies either correct a specific axis such as length~\citep{park2024disentangling,lu2024sampo} or add hand-constructed pairs whose content is equal and only surface features differ~\citep{moya2026spurious}. In the self-generated pairs described above, this problem takes its strongest form. The chosen (gold) and rejected (the model's own) responses differ not only in length but in \emph{style} in general---format, verbosity, phrasing. Moreover, when the target knowledge is partially known to the model, most rejected responses are factually \emph{correct}, so that pairs differing only in style and pairs with a genuine factual difference are mixed together without distinction. Standard DPO makes no such distinction and pushes down the whole rejected response on every pair, and with the style it pushes down the correct knowledge CPT has just acquired. Our work differs from methods that correct a specific axis such as length~\citep{park2024disentangling,lu2024sampo} in removing the influence of style as a whole. The method that adds pairs whose content is equal and only surface features differ~\citep{moya2026spurious} shares with ours the treatment of style as a whole, but it constructs such pairs by hand, adds them to the training data, and trains on them with random labels; we instead identify such pairs, which are already present in self-generated data, with a judge LLM, and label them not at random but in the direction and with the weight that exactly cancel the style signal of the pairs with a factual difference.

Our goal is to remove from these preference pairs the learning signal that stems from differences in style, and to update the model with the signal that stems from differences in fact alone. To this end we propose \emph{style-debiased DPO} (SD-DPO). For each pair, a judge LLM scores whether the rejected response agrees with the chosen one on the facts, and the pairs are split into a group with a genuine factual difference and a group that is factually correct and differs only in style. The former is trained as usual; the latter is trained with its preference \emph{inverted}. The inverted group receives a weight chosen so that the gradients due to differences in style (henceforth the \emph{style gradients}) of the two groups balance. Under mild assumptions, the style component vanishes from the expected gradient at initialization and only the factual learning signal remains (Proposition~\ref{prop:cancel}). The assumption stated above is supported as follows. The CPT model's responses agree with the gold answer on the facts in about two thirds of the preference data (Section~\ref{sec:exp_setup}), so storage has taken place. On top of that, with the same gold answers as chosen responses, standard DPO, which pushes every rejected response down, does not improve accuracy, whereas SD-DPO, which does not push down the factually correct ones, improves it (Section~\ref{sec:results_quality}) while preserving general capabilities (Section~\ref{sec:results_general}). This contrast indicates that the gain comes from how knowledge is elicited, not from new knowledge.

Our contributions are threefold. \textbf{(i)}~We identify and quantify how self-generated preference data fails for knowledge updating---factually correct rejected responses dominate whenever the target knowledge is partially known---and propose SD-DPO, for which we prove that the preference inversion cancels the style gradient at initialization. \textbf{(ii)}~On the reading-comprehension QA benchmark QuALITY, we exceed a CPT baseline trained in the EntiGraph setting and evaluated under the same conditions as our method; the preference data this requires is a few percent of the additional CPT tokens that would be needed to reach the same accuracy by CPT alone. \textbf{(iii)}~For knowledge updating, the main goal of this work, on the knowledge-editing benchmark AToKE we attain high new-fact acquisition and old-fact recall simultaneously, which methods that rewrite model parameters directly do not (reference comparison; different protocol).

\section{Related Work}
\label{sec:related}

\textbf{Storing knowledge with CPT on synthetic data.}
\citet{allen2024physics} showed, with models trained from scratch, that the diversity of paraphrases in the training data governs whether facts can later be extracted. \citet{yang2025entigraph}, with EntiGraph, extended this insight to CPT of pretrained LLMs and showed on QuALITY~\citep{pang2022quality}, a reading-comprehension QA dataset over long stories, that accuracy grows with the number of synthetic tokens following a mixture-of-exponentials scaling law. \citet{cheng2024adapting} convert raw text into reading-comprehension format for training. All of these optimize storage; none improves retrieval with a preference stage after CPT.

\textbf{Updating knowledge with preference optimization.}
PASTA~\citep{yamamoto2026pasta} applies CPT, supervised fine-tuning (SFT), and Self-DPO in sequence to news articles past the knowledge cutoff, and \citet{rozner2024knowledge} cast the editing of a single fact as DPO with the old fact as rejected. In both, most rejected responses are factually wrong, so standard DPO receives an unambiguous learning signal. Our setting, where the target knowledge is partially known and self-generated rejected responses are mostly \emph{correct}, has, to our knowledge, not been addressed explicitly.

\textbf{Noise-robust preference optimization.}
rDPO~\citep{chowdhury2024provably}, $\beta$-DPO~\citep{wu2024beta}, and filtered DPO~\citep{morimura2024filtered} treat imperfect pairs as label noise or low quality; none asks \emph{what} differs within a pair, although reward models are easily misled by format over content~\citep{liu2024rmbench}, preference tuning amplifies style over content features~\citep{ferrao2025anatomy}, and DPO gains little signal when chosen and rejected distributions are similar~\citep{pan2025whatmatters}, as here.

\textbf{Length bias and spurious features.}
The remedies for the problem described in Section~\ref{sec:intro}, that features unrelated to content (spurious features) such as length are learned, are an explicit length margin~\citep{park2024disentangling}, SamPO, which computes the reward over an equal number of tokens drawn from both responses~\citep{lu2024sampo}, and \emph{tie training}, which adds pairs of equal utility that differ only in spurious features, labeled at random~\citep{moya2026spurious}. Our style-only pairs are such ties, found automatically by the judge LLM's measure of factual agreement (the factual-agreement score defined in Section~\ref{sec:fa}), and SD-DPO labels them so that the mean of the spurious feature cancels rather than at random (Appendix~\ref{app:ties}); because our pairs differ greatly in length (Appendix~\ref{app:length}), we also compare against SamPO under identical conditions.

\textbf{Knowledge editing.}
ROME~\citep{meng2022locating} and MEMIT~\citep{meng2023memit} write individual facts into parameters with high efficacy, but on knowledge that changes over time, editing overwrites the old fact. \citet{yin2024history} propose AToKE, a knowledge-editing benchmark for updating facts that change over time, and show that models edited with standard ROME or MEMIT answer questions about the old fact for its period correctly only about 2\% of the time. The same paper also proposes METO, which modifies an existing editing method so that it edits the new and the old fact together with the period in which each holds; applied to ROME and MEMIT, METO raises this accuracy to 20--30\%, still far below the new-fact accuracy the same paper reports for these editors (86--100\%). We approach AToKE through CPT plus preference optimization and retain both facts.

\begin{figure}[!t]
\centering
\includegraphics[width=\linewidth]{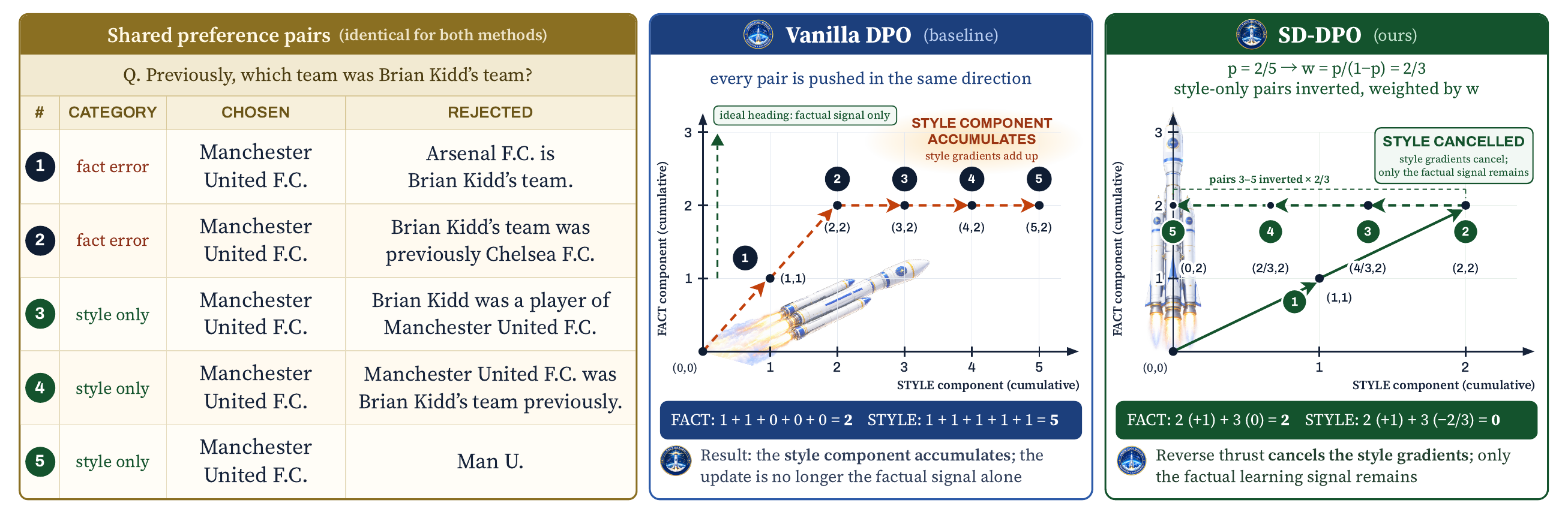}
\caption{Conceptual illustration of SD-DPO: the learning signal in the style direction is cancelled so that only the factual (upward) component is learned, drawn as a rocket launching straight up. Both methods share five pairs (two with a factual difference, three style-only); each pair is one burst of thrust, upward for the factual component and sideways for the style component, and the axes are cumulative. Vanilla DPO learns all five in the same direction and ends at (style 5, fact 2). SD-DPO inverts the style-only pairs (reverse thrust) with weight $w=p/(1-p)=2/3$, so the style components cancel, $2\,(+1)+3\,(-2/3)=0$, and the end point is (style 0, fact 2).}
\label{fig:method}
\end{figure}

\section{Method}
\label{sec:method}

We first give an overview of the method described in this section. The input is a source corpus. The goal is to train the target LLM on this knowledge, so that afterwards it answers a variety of questions about that knowledge correctly without seeing the source text. The method has two stages. In the first stage we generate synthetic data from the source corpus and run CPT on it; the synthetic data are texts that explain the relations among the entities in a document in many different phrasings, as in EntiGraph, or paraphrases of the sentence stating each fact together with QA about its content. After this stage the knowledge is stored in the model, but it is not necessarily elicited correctly when the model is asked. The second stage is the proposed SD-DPO. We build preference pairs whose chosen response is the gold answer of a QA built from the source text and whose rejected response is what the CPT model itself returns to the same question. DPO would then suppress the rejected response and reinforce the chosen one. However, as described in Section~\ref{sec:intro}, most rejected responses are factually correct and differ from the chosen ones only in style. In SD-DPO, a judge LLM scores the factual agreement of each pair; only pairs whose rejected response is factually wrong are trained in the usual DPO direction, while pairs that differ only in style are trained with the direction inverted, with a weight chosen so that the style-related gradients of the two groups exactly balance. As a result, the style signal cancels over the whole dataset and only the factual content is learned.

\subsection{Problem setup and notation}
\label{sec:setup}

Let $\mathcal{C}$ be the source corpus and $\mathcal{Q}=\{(q_i, a_i^{\text{gold}})\}$ the question--answer pairs built from it. We call a model that defines a probability distribution over responses to a question a \emph{policy}. The model obtained by CPT on a synthetic corpus derived from $\mathcal{C}$ is the reference policy $\piref$, which is also the initialization of the policy $\pi_\theta$ being trained. The preference data is $\mathcal{D}=\{(q_i, y_i^w, y_i^l)\}_{i=1}^N$, where the chosen response $y_i^w$ is the gold answer $a_i^{\text{gold}}$ and the rejected response $y_i^l\sim\piref(\cdot\mid q_i)$ is sampled from the reference policy. The \emph{margin} measures how far training has moved the log-likelihood gap between chosen and rejected away from its value under the reference policy,
\begin{equation}
\Delta_i \;=\; \log\tfrac{\pi_\theta(y_i^w\mid q_i)}{\pi_\theta(y_i^l\mid q_i)} \;-\; \log\tfrac{\piref(y_i^w\mid q_i)}{\piref(y_i^l\mid q_i)},
\end{equation}
and the standard DPO loss is $\mathcal{L}_{\text{DPO}}=\mathbb{E}_i[-\log\sigma(\beta\Delta_i)]$, where $\sigma$ is the sigmoid function and $\beta$ controls how far the policy may move from the reference. At initialization $\pi_\theta=\piref$, so $\Delta_i=0$, $\sigma(0)=1/2$, and the gradient reduces to
\begin{equation}
\label{eq:grad}
\nabla_\theta\mathcal{L}_{\text{DPO}}\big|_{\theta_{\text{ref}}} = -\tfrac{\beta}{2}\,\mathbb{E}_i\big[\nabla_\theta\log\pi_\theta(y_i^w\mid q_i)-\nabla_\theta\log\pi_\theta(y_i^l\mid q_i)\big].
\end{equation}
The second term \emph{directly lowers} the likelihood of the rejected response---the operation we want when $y^l$ is wrong, and the operation we must avoid when $y^l$ is right. DPO also solves the same optimization problem as reinforcement learning from human feedback (RLHF), whose solution $\pi^\star$ cannot move far from the reference policy: in terms of the KL divergence, a measure of the gap between two distributions, $\mathrm{KL}(\pi^\star\|\piref)\le\Delta r/\beta$, where $\Delta r$ is the reward gained by the optimization. Under a mild assumption, how much the knowledge stored by CPT can change is bounded from above by this KL (Appendix~\ref{app:theory}). This is why a preference stage can change \emph{how knowledge is accessed} while approximately preserving \emph{what is stored}---and why what endangers stored knowledge is training far past the optimum, not the shape of the loss (Section~\ref{sec:budget}).

\subsection{Measuring factual agreement}
\label{sec:fa}

A judge LLM $J$ scores, for each pair, how far the rejected response agrees with the chosen one on the facts, on a scale from 0 to 1. We call this score the \emph{factual agreement} and compute it in two ways. The \emph{pairwise} score $f_i^{(\mathrm{P})}=J(q_i,y_i^w,y_i^l)\in[0,1]$ compares the two responses using only the question and the responses themselves. The \emph{context-grounded} score also gives the judge the document $d_i$ from which the question was built, and asks it to score the rejected response for factual accuracy against the document rather than for agreement with the chosen response, which also catches errors in details the chosen response omits. This score is multiplied by a grounding factor measuring how much a response draws on the document,
\begin{equation}
f_i^{(\mathrm{C})} \;=\; J(d_i, q_i, y_i^w, y_i^l)\cdot\min\!\big(1,\; v_i^l/v_i^w\big),
\end{equation}
where $v^w$ and $v^l$ are the fractions of words in the chosen and rejected responses that fall inside spans of three or more consecutive words matching $d_i$; the factor discounts generic answers that never touch the document's content. We call the pairwise score pFA and the context-grounded score cFA. The choice depends on whether the rejected response's correctness can be decided from the chosen response alone. For questions about a single fact, such as a person or a date, we use pFA. For questions about a character's motivation or the causes of events, where the rejected response gives reasons different from the chosen one and correctness is decidable only against the source, we use cFA. Let $p$ be the share of pairs with factual agreement below $0.5$, i.e., a wrong rejected response.

\subsection{Style-debiased DPO}
\label{sec:sddpo}

We assume that the margin $\Delta_i$ is approximately the sum of a component due to the factual difference between chosen and rejected, $\Delta_i^{(\mathrm{F})}$, and a component due to their difference in style, $\Delta^{(\mathrm{S})}$: $\Delta_i\approx\Delta_i^{(\mathrm{F})}+\Delta^{(\mathrm{S})}$; we call this approximation the \emph{additive decomposition} of the margin. The style component carries no pair index $i$ because we assume it is shared across pairs: all chosen responses follow the same template and all rejected responses are generated by the same model (Assumption A2; the case where it fails is analyzed in Appendix~\ref{app:theory}). Let $c_i=1$ for pairs whose factual agreement is below the threshold $\thr=0.5$, i.e., whose rejected response is factually wrong, and $c_i=0$ for the remaining style-only pairs ($c_i=\mathds{1}[f_i<\thr]$), and let $p=\Pr[c_i=1]$ be the share of $c_i=1$ pairs. SD-DPO keeps the DPO training direction on pairs with a factual difference, inverts it on style-only pairs, and multiplies the inverted group by the weight $w=p/(1-p)$:
\begin{equation}
\label{eq:sddpo}
\mathcal{L}_{\text{SD}} \;=\; c_i\big[-\log\sigma(\beta\Delta_i)\big] \;+\; (1-c_i)\,w\,\big[-\log\sigma(-\beta\Delta_i)\big].
\end{equation}
Dropping or down-weighting style-only pairs cannot remove the style gradient of the rest; cancellation requires an opposing gradient of matched magnitude, which the inversion and $w$ provide.

\begin{proposition}[Style-gradient cancellation at initialization]
\label{prop:cancel}
Under the additive decomposition and Assumption A2, with $N$ pairs, the expected gradient of \eqref{eq:sddpo} at initialization $\theta=\theta_{\text{ref}}$ is proportional to
$-\big[\sum_{i:c_i=1}\nabla_\theta\Delta_i^{(\mathrm{F})}-w\sum_{i:c_i=0}\nabla_\theta\Delta_i^{(\mathrm{F})}\big]$: the style terms sum to $Np - w\,N(1-p) = 0$ and vanish for any $p\in(0,1)$.
\end{proposition}

The weight $w$ is determined uniquely by this cancellation condition and is not tuned. The guarantee concerns the gradient averaged over the whole dataset at initialization; we keep the model near that regime by averaging the gradients of many pairs per step and by the budget control of Section~\ref{sec:budget}.

\Eqref{eq:sddpo} is the ideal form that the analysis addresses. The loss we actually train makes three changes to it. First, each pair is multiplied by a weight $\hat{\phi}_i$ that reflects how confident the judge score is, normalized to mean one within each group. Second, the two group weights $(1,\,w)$ are rescaled to $(1-p,\,p)$, preserving their ratio. Third, the term of the inverted group is augmented with a knowledge-preservation term, a hinge penalty that keeps the likelihood of the correct rejected response from falling below its value under the reference policy:
\begin{align}\label{eq:impl}
\mathcal{L}^{\text{impl}}_{\text{SD}}
&= c_i(1{-}p)\,\hat{\phi}_i\big[{-}\log\sigma(\beta\Delta_i)\big] \nonumber\\
&\quad + (1{-}c_i)\,p\,\hat{\phi}_i\Big[{-}\log\sigma({-}\beta\Delta_i)
+ \lambda_{\text{KP}}\max\!\big(0,\log\piref(y_i^l\mid q_i)-\log\pi_\theta(y_i^l\mid q_i)\big)\Big],
\end{align}
The knowledge-preservation coefficient is $\lambda_{\text{KP}}=1$. Proposition~\ref{prop:cancel} still holds for \eqref{eq:impl}: since $\hat{\phi}_i$ has mean one within each group, the cancellation holds in expectation, and the knowledge-preservation term is inactive at initialization, where the argument of the $\max$ is exactly zero. Replay data (general-purpose preference pairs added to prevent forgetting) is trained with standard DPO, and the end-of-sequence (EOS) token is excluded from the likelihood sums that define $\Delta_i$: without this exclusion, lowering chosen likelihoods on the inverted group, which is the majority, also suppresses how every chosen response ends, and the model lapses into endless repetition (Appendix~\ref{app:impl}).

\section{Experiments}
\label{sec:experiments}

This section tests whether the proposed method improves the elicitation of knowledge stored by CPT, on two benchmarks of different character. We choose the reading-comprehension QA dataset QuALITY because EntiGraph, a representative method on the storing side, was evaluated on it, so we can test on the same synthetic data and with the same evaluation procedure whether learning on the eliciting side adds accuracy on top of its CPT model. Here we measure the accuracy gain over the CPT baseline and the amount of preference data it requires. Comparing standard DPO and its variants (IPO, KTO, SamPO) under identical conditions isolates the contribution of the preference inversion (Section~\ref{sec:results_quality}). Knowledge updating, the main goal of this work, is tested on the knowledge-editing benchmark AToKE. We choose AToKE because it asks, for knowledge in which an old and a new fact coexist, whether the model retrieves the fact that holds at the queried period, and thus directly measures what knowledge updating requires: acquiring the new fact while keeping the old one (Section~\ref{sec:results_atoke}). Finally, we check on six standard benchmarks that this training does not harm general capabilities (Section~\ref{sec:results_general}). The setup is given in Section~\ref{sec:exp_setup} and additional results in Appendix~\ref{app:extra}.

\subsection{Setup}
\label{sec:exp_setup}

\textbf{QuALITY (elicitation efficiency).} Following EntiGraph, we CPT Llama-3-8B-Instruct~\citep{dubey2024llama} on a 449M-token synthetic corpus built from the 265 QuALITY articles, then apply preference optimization to 52{,}582 pairs (21.7M tokens; both multiple-choice and free-form questions) built from QA that follows chains of entity relations in the source articles. Factual agreement uses cFA because rejecting the distractors requires article-specific content. The share of pairs whose rejected response is factually wrong ($p$ of Section~\ref{sec:sddpo}) is about 0.33, so the remaining two thirds differ only in style. Evaluation is closed-book 4-way multiple choice (answering from memory, without the article) on all 4{,}609 QuALITY questions about these articles, as in EntiGraph (Appendix~\ref{app:hyper}). We sample $n{=}8$ responses per question and score one drawn uniformly at random from those that follow the answer format (``random valid''); its expected accuracy is independent of $n$. The baseline is our CPT model, which is also the starting point of SD-DPO, evaluated with the same procedure (code, prompts, sampling) as our method: 56.21\% (differences from EntiGraph's setup: Appendix~\ref{app:hyper}). As in EntiGraph, QuALITY uses this single model; robustness across model families is tested on AToKE, our main goal.

\textbf{AToKE (knowledge updating).} Our main base model is Llama-3.1-8B-Instruct; to check that the results do not depend on the model, we apply the same procedure to Qwen3-8B and Gemma4-E2B (Appendix~\ref{app:cross}). Numbers below refer to Llama unless stated otherwise. We use 8{,}819 edit cases; each has an old and a new fact about a subject, with the period each was true. For each case we synthesize four seed sentences (new and old fact, each with an explicit and with a relative time), paraphrase each 10 times, and generate one QA per sentence. CPT (19.7M tokens) is followed by SD-DPO on 387{,}287 pairs. Factual agreement uses pFA, because there is no long document ($p=0.148$, i.e., 85.2\% of pairs differ only in style). Four metrics: CES and CRS are accuracies on new-fact questions with and without an explicit period, HES and HRS likewise for the old fact; Current (new-fact acquisition), History (old-fact recall), and Overall are the means of (CES, CRS), (HES, HRS), and all four. \texttt{gpt-5-mini} judges factual agreement and scores AToKE; hyperparameters are in Appendix~\ref{app:hyper}.

\paragraph{Controlling the training budget.}\label{sec:budget}
Training on all preference data at the DPO paper's learning rate, $1{\times}10^{-6}$, sharply drops accuracy late in training for every loss we tried (standard DPO, SD-DPO, and the DPO variants IPO~\citep{azar2024general} and KTO~\citep{ethayarajh2024kto}), consistent with reward over-optimization~\citep{rafailov2024scaling}. With a constant learning rate, no warmup, and one epoch, the total parameter update is proportional to learning rate ${\times}$ steps, so early stopping, filtering out pairs with factual agreement at least $\fmax$ (whose rejected response is correct), and lowering the learning rate are interchangeable. We mainly use $5{\times}10^{-7}$, the common choice for DPO on 7--8B models. Combining SD-DPO with the $\fmax$ filter leaves only $c_i=1$ pairs, so nothing is inverted; Section~\ref{sec:results_quality} uses this as an ablation removing only the inversion (mechanism ablation).

\subsection{QuALITY: exceeding the CPT baseline efficiently}
\label{sec:results_quality}

\begin{table}[t]
\centering
\caption{QuALITY 4-way multiple-choice accuracy (\%; same evaluation procedure for all rows; seed 42). Each preference-optimization row shows its best checkpoint, with its usage (the share of preference data trained on) in parentheses; ``$+$tokens'' is the preference data consumed.}
\label{tab:quality_main}
\small
\begin{tabular}{llrcc}
\toprule
Stage & Configuration & $+$tokens & MCQ (\%) & $\Delta$ \\
\midrule
Llama-3-8B-Instruct & --- & --- & 41.24 & --- \\
$+$ EntiGraph CPT (\textbf{baseline}) & 449M synthetic corpus & --- & 56.21 & $+14.97$ \\
\midrule
$+$ Vanilla DPO (original setup) & unfiltered, lr $10^{-6}$ (25\%) & 5.4M & 52.77 & $-3.44$ \\
$+$ Vanilla DPO (best config.) & pFA, $\fmax{=}0.5$, lr $10^{-6}$ (10\%) & 0.26M & 56.45 & $+0.24$ \\
$+$ IPO & unfiltered, lr $5{\times}10^{-7}$ (25\%) & 5.4M & 55.68 & $-0.53$ \\
$+$ KTO & unfiltered, lr $5{\times}10^{-7}$ (10\%) & 2.2M & 56.05 & $-0.16$ \\
$+$ SamPO (length-debiased) & unfiltered, lr $5{\times}10^{-7}$ (100\%) & 21.8M & 57.32 & $+1.11$ \\
$+$ \textbf{SD-DPO (ours)} & cFA, unfiltered, lr $5{\times}10^{-7}$ (50\%) & 10.9M & \textbf{59.17} & $\bm{+2.96}$ \\
\bottomrule
\end{tabular}
\end{table}

\begin{table}[t]
\centering
\caption{Best\,/\,final accuracy (\%) and drop (final $-$ best) under three learning rates (no filter) and the $\fmax$ filter (learning rate $10^{-6}$). $\dagger$: mechanism ablation (inversion removed; Section~\ref{sec:budget}), not a recommended setting.}
\label{tab:quality_lr}
\small
\setlength{\tabcolsep}{5pt}
\begin{tabular}{lccccc}
\toprule
 & \multicolumn{3}{c}{Learning rate (unfiltered)} & \multicolumn{2}{c}{$\fmax{=}0.5$ (lr $10^{-6}$)} \\
\cmidrule(lr){2-4}\cmidrule(lr){5-6}
 & $10^{-6}$ & $5{\times}10^{-7}$ & $2.5{\times}10^{-7}$ & pFA & cFA \\
\midrule
SD-DPO (cFA) & 56.61\,/\,50.56 & \textbf{59.17\,/\,57.98} & 58.01\,/\,57.22 & --- & 55.82\,/\,53.30$^{\dagger}$ \\
\quad drop & $-6.05$ & $\bm{-1.19}$ & $-0.79$ & --- & $-2.52$ \\
\midrule
Vanilla DPO & 52.77\,/\,47.01 & 56.01\,/\,54.64 & 56.53\,/\,56.27 & 56.45\,/\,55.99 & 55.29\,/\,48.84 \\
\quad drop & $-5.76$ & $-1.37$ & $-0.26$ & $-0.46$ & $-6.45$ \\
\bottomrule
\end{tabular}
\end{table}

Table~\ref{tab:quality_main} shows the main result. SD-DPO reaches 59.17\% after training on 50\% of the preference data (``usage 50\%''), $+2.96$ over the baseline, and still 57.98\% ($+1.77$) at the final checkpoint after all of it. Vanilla DPO with the original DPO paper's settings (learning rate $10^{-6}$, no filter) is below the baseline at every checkpoint (best 52.77\%), confirming the failure described in Section~\ref{sec:intro}, and $\fmax$ filtering without inversion (Vanilla $+\fmax$ in the table) stays at the baseline level ($+0.24$, within the seed-to-seed standard deviation of 1.75). The improvement survives a change of seed: the best checkpoints of seeds 42/43/44 average $58.55\pm1.75$, all above baseline (Appendix~\ref{app:extra}).

Table~\ref{tab:quality_lr} separates the effects of the two ingredients, inversion and training budget. \textbf{(a) The late-training drop stems from the training budget, not from the loss:} at learning rate $10^{-6}$, both vanilla DPO and SD-DPO end about 6 points below their best checkpoint; halving the learning rate shrinks the drop to $-1.19$ while \emph{raising} the best accuracy to 59.17\%. \textbf{(b) The gain comes from the inversion, not from tuning the budget:} at every learning rate vanilla DPO stays within noise of the baseline (best 56.53\% at $2.5{\times}10^{-7}$), and none of nine filter-strength/learning-rate combinations exceeds $+0.58$ (Appendix~\ref{app:extra}). \textbf{(c) Removing only the inversion (mechanism ablation;} $\dagger$ in the table\textbf{)} drops SD-DPO from 59.17\% to 55.82\%. The same conclusion holds for the DPO variants IPO and KTO trained under identical conditions: IPO regresses the preference margin toward a finite target instead of increasing it indefinitely, which changes how \emph{hard} rejected responses are pushed down but not, near initialization, the \emph{direction}, and KTO treats chosen and rejected responses individually rather than as pairs but still treats correct rejected responses as undesirable; neither exceeds the baseline. \textbf{(d) Length debiasing alone helps, but is not the main source of the gain:} SamPO, which computes the reward over an equal number of token positions drawn at random from both responses (the shorter one's count) and so removes the influence of length (Appendix~\ref{app:extra}), reaches 57.32\% ($+1.11$)---above the baseline, confirming that part of vanilla DPO's failure is the length asymmetry of our pairs---yet stays 1.85 points below SD-DPO. The two improve accuracy in opposite ways: SamPO's outputs shrink to the bare ``\texttt{Answer: X.}'' (2 words; interpretation in Appendix~\ref{app:extra}), whereas SD-DPO's contain the recalled grounds (200 words).

\begin{figure}[t]
\centering
\includegraphics[width=0.74\linewidth]{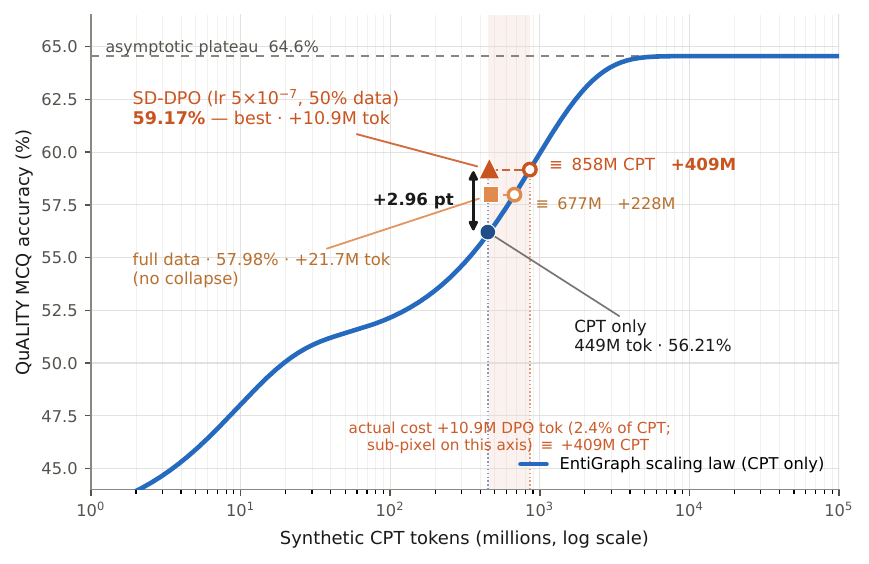}
\caption{The relation between training tokens and accuracy reported for EntiGraph (its scaling law), with our CPT model and SD-DPO placed on it.}
\label{fig:scaling}
\end{figure}

Figure~\ref{fig:scaling} converts the gain into an equivalent number of CPT tokens. Since our CPT model's accuracy falls on the tokens-versus-accuracy curve reported for EntiGraph (56.1\% predicted, 56.21\% measured), we can ask what additional CPT budget would reach 59.17\%: about $409$M tokens, forty times the preference data we used. In this regime, eliciting stored knowledge is far cheaper than storing more, though this is a single measured point and we did not map where the trade-off reverses.

\subsection{AToKE: retrieving the fact that holds at the queried period}
\label{sec:results_atoke}

\begin{table}[t]
\centering
\caption{The single-edit setting of AToKE (Single Edit; accuracy 0--1). CES/CRS query the new fact (with/without explicit period); HES/HRS query the old fact. Rows above the rule: parameter editing on GPT-J~6B from \citet{yin2024history} (different base and protocol; reference only; ``$+$METO'' = with METO). Our rows share data, base model, and evaluation procedure and are directly comparable with one another.}
\label{tab:atoke}
\footnotesize
\setlength{\tabcolsep}{5pt}
\begin{tabular}{lccccccc}
\toprule
Method & CES & HES & CRS & HRS & Current & History & Overall \\
\midrule
ROME & \textbf{1.000} & 0.024 & 0.816 & 0.016 & 0.908 & 0.020 & 0.464 \\
MEMIT & 0.997 & 0.022 & 0.753 & 0.012 & 0.875 & 0.017 & 0.446 \\
MEND & 0.805 & 0.017 & 0.325 & 0.007 & 0.565 & 0.012 & 0.288 \\
ROME$+$METO & \textbf{1.000} & 0.203 & 0.789 & 0.163 & 0.894 & 0.183 & 0.538 \\
MEMIT$+$METO & 0.864 & 0.303 & 0.741 & 0.243 & 0.802 & 0.273 & 0.538 \\
\midrule
Base (before CPT) & 0.038 & 0.179 & 0.011 & 0.115 & 0.025 & 0.147 & 0.086 \\
CPT only & 0.905 & 0.782 & 0.880 & 0.974 & 0.892 & 0.878 & 0.885 \\
$+$ Vanilla DPO & 0.608 & 0.823 & 0.633 & 0.887 & 0.620 & 0.855 & 0.738 \\
$+$ SD-DPO (ours) & 0.985 & \textbf{0.954} & \textbf{0.989} & \textbf{0.998} & \textbf{0.987} & \textbf{0.976} & \textbf{0.982} \\
\bottomrule
\end{tabular}
\end{table}

\begin{figure}[t]
\centering
\definecolor{mLlama}{HTML}{0072B2}
\definecolor{mQwen}{HTML}{E69F00}
\definecolor{mGemma}{HTML}{009E73}
\definecolor{starzone}{HTML}{C8F000}
\begin{tikzpicture}
\begin{axis}[
  width=\linewidth, height=7.6cm,
  xlabel={History (old-fact recall)},
  ylabel={Current (new-fact acquisition)},
  xmin=-0.02, xmax=1.02, ymin=0.33, ymax=1.03,
  xtick={0,0.2,0.4,0.6,0.8,1.0}, ytick={0.4,0.6,0.8,1.0},
  grid=major, grid style={dashed,gray!20},
  tick label style={font=\footnotesize}, label style={font=\footnotesize},
  xlabel style={yshift=4pt}, ylabel style={yshift=-1pt},
  clip=false,
  legend style={at={(0.025,0.04)}, anchor=south west, draw=none, fill=none, font=\scriptsize,
    legend columns=2, column sep=14pt, row sep=2pt, /tikz/every even column/.append style={column sep=6pt}},
  legend cell align=left, legend image post style={mark size=2.6},
]
\fill[starzone!30] (axis cs:0.77,1.03) -- (axis cs:1.02,0.78) -- (axis cs:1.02,1.03) -- cycle;
\addplot[forget plot, domain=0.00:0.68, dashed, gray!55, thin] {1.0-x}
  node[pos=0.30, sloped, below, font=\scriptsize, gray!75!black] {Overall $=0.5$};
\addplot[forget plot, domain=0.38:0.88, dashed, gray!55, thin] {1.4-x}
  node[pos=0.20, sloped, above, font=\scriptsize, gray!75!black] {$0.7$};
\addplot[forget plot, domain=0.77:1.02, dashed, gray!55, thin] {1.8-x}
  node[pos=0.88, sloped, above, font=\scriptsize, gray!75!black] {$0.9$};
\addplot[forget plot, only marks, mark=diamond, mark size=2.8, gray!55!black] coordinates {
  (0.020,0.908) (0.017,0.875) (0.012,0.565) (0.183,0.894) (0.273,0.802)};
\node[font=\scriptsize, gray!55!black, anchor=west] at (axis cs:0.030,0.925) {ROME};
\node[font=\scriptsize, gray!55!black, anchor=west] at (axis cs:0.030,0.858) {MEMIT};
\node[font=\scriptsize, gray!55!black, anchor=west] at (axis cs:0.025,0.565) {MEND};
\node[font=\scriptsize, gray!55!black, anchor=west] at (axis cs:0.198,0.894) {ROME$+$METO};
\node[font=\scriptsize, gray!55!black, anchor=west] at (axis cs:0.288,0.802) {MEMIT$+$METO};
\draw[-{Latex[length=1.8mm]}, mLlama!60, semithick] (axis cs:0.893,0.905) -- (axis cs:0.963,0.975);
\draw[-{Latex[length=1.8mm]}, mQwen!60, semithick] (axis cs:0.848,0.860) -- (axis cs:0.886,0.876);
\draw[-{Latex[length=1.8mm]}, mGemma!60, semithick] (axis cs:0.702,0.704) -- (axis cs:0.900,0.900);
\draw[-{Latex[length=1.8mm]}, mLlama!60, semithick, dashed] (axis cs:0.875,0.872) -- (axis cs:0.858,0.642);
\draw[-{Latex[length=1.8mm]}, mQwen!60, semithick, dashed] (axis cs:0.826,0.836) -- (axis cs:0.620,0.388);
\draw[-{Latex[length=1.8mm]}, mGemma!60, semithick, dashed] (axis cs:0.690,0.673) -- (axis cs:0.711,0.641);
\addplot[forget plot, only marks, mark=o, mark size=2.6, very thick, mLlama] coordinates {(0.878,0.892)};
\addplot[forget plot, only marks, mark=o, mark size=2.6, very thick, mQwen] coordinates {(0.836,0.853)};
\addplot[forget plot, only marks, mark=o, mark size=2.6, very thick, mGemma] coordinates {(0.683,0.685)};
\addplot[forget plot, only marks, mark=triangle*, mark size=3.0, mLlama, mark options={fill=mLlama, draw=mLlama!65!black, line width=0.25pt}] coordinates {(0.855,0.620)};
\addplot[forget plot, only marks, mark=triangle*, mark size=3.0, mQwen, mark options={fill=mQwen, draw=mQwen!65!black, line width=0.25pt}] coordinates {(0.608,0.362)};
\addplot[forget plot, only marks, mark=triangle*, mark size=3.0, mGemma, mark options={fill=mGemma, draw=mGemma!65!black, line width=0.25pt}] coordinates {(0.716,0.629)};
\node[star, star points=5, star point ratio=2.3, inner sep=1.55pt, fill=mLlama, draw=mLlama!70!black, line width=0.25pt] at (axis cs:0.976,0.987) {};
\node[star, star points=5, star point ratio=2.3, inner sep=1.55pt, fill=mQwen, draw=mQwen!70!black, line width=0.25pt] at (axis cs:0.899,0.881) {};
\node[star, star points=5, star point ratio=2.3, inner sep=1.55pt, fill=mGemma, draw=mGemma!70!black, line width=0.25pt] at (axis cs:0.916,0.916) {};
\draw[-{Latex[length=2.2mm]}, very thick, black!55]
  (axis cs:0.760,0.375) -- (axis cs:0.860,0.475);
\node[font=\footnotesize\bfseries, anchor=south, black!60] at (axis cs:0.835,0.490) {top right is better};
\addlegendimage{only marks, mark=o, mark size=2.6, very thick, gray!45!black}\addlegendentry{\textcolor{gray!45!black}{CPT only}}
\addlegendimage{only marks, mark=square*, mark size=2.4, mLlama, mark options={fill=mLlama, draw=mLlama!65!black, line width=0.25pt}}\addlegendentry{\textcolor{mLlama}{Llama-3.1-8B}}
\addlegendimage{only marks, mark=triangle*, mark size=3.0, gray!45!black, mark options={fill=gray!45!black, draw=black!60, line width=0.25pt}}\addlegendentry{\textcolor{gray!45!black}{$+$ vanilla DPO}}
\addlegendimage{only marks, mark=square*, mark size=2.4, mQwen, mark options={fill=mQwen, draw=mQwen!65!black, line width=0.25pt}}\addlegendentry{\textcolor{mQwen}{Qwen3-8B}}
\addlegendimage{legend image code/.code={\node[star, star points=5, star point ratio=2.3, inner sep=0pt, minimum size=6.6pt, fill=gray!45!black, draw=black!60, line width=0.25pt] at (0.223cm,+0.46pt) {};}}\addlegendentry{\textcolor{gray!25!black}{\bfseries $+$ SD-DPO (ours)}}
\addlegendimage{only marks, mark=square*, mark size=2.4, mGemma, mark options={fill=mGemma, draw=mGemma!65!black, line width=0.25pt}}\addlegendentry{\textcolor{mGemma}{Gemma4-E2B}}
\end{axis}
\end{tikzpicture}
\vspace{-15pt}
\caption{New-fact acquisition (Current) vs.\ old-fact recall (History) on AToKE. Colour is the base model and shape the method (CPT only $\circ$, vanilla DPO $\triangle$, SD-DPO $\star$); arrows leave the CPT model, solid to SD-DPO and dashed to vanilla DPO. Grey dashed lines join equal Overall, the shaded band is Overall $\geq 0.9$, and open diamonds are prior parameter edits (Table~\ref{tab:atoke}). Values: Tables~\ref{tab:atoke} and~\ref{tab:atoke_cross_full}.}
\label{fig:atoke_tradeoff}
\vspace{-9pt}
\end{figure}
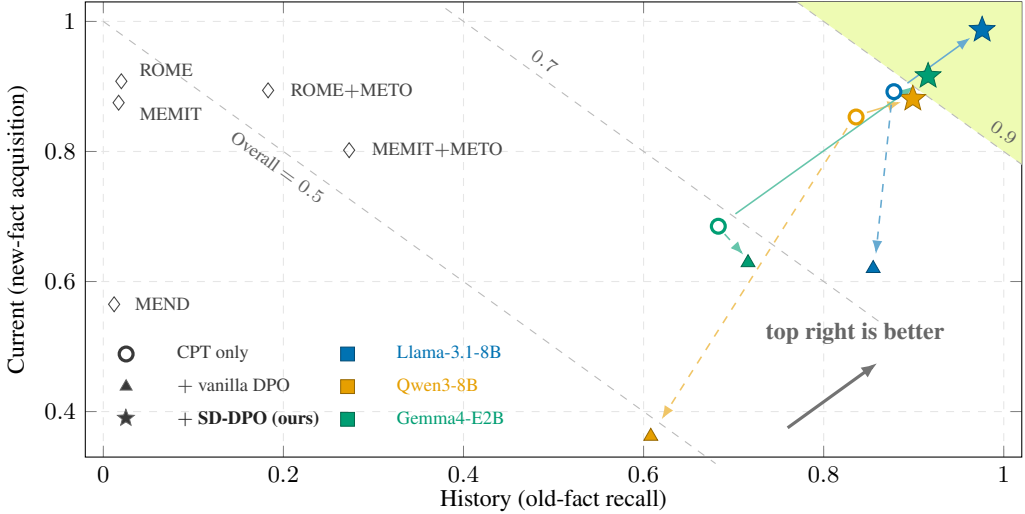

Table~\ref{tab:atoke} evaluates updating knowledge whose answer changes with time: the old fact is not an error but the answer for an earlier period. As old-fact QA is in the CPT corpus and the preference data, HES/HRS measure whether the model \emph{produces the fact that holds at the queried period}, not mere retention. SD-DPO attains 0.982 Overall; HES (old fact, explicit period) rises from 0.782 to 0.954, whereas parameter editing reaches at most 0.303 (Table~\ref{tab:atoke}, top). Swapping only the loss for standard DPO \emph{reverses} the outcome: because most pairs differ only in style, vanilla DPO pushes down knowledge CPT just acquired and Overall falls to 0.738, \emph{below} CPT only (Figure~\ref{fig:atoke_tradeoff})---the clearest evidence that the inversion is decisive. We observe the same contrast on two further base models of different families and sizes, Qwen3-8B and Gemma4-E2B: on both, SD-DPO exceeds CPT only and vanilla DPO, and vanilla DPO falls below CPT only (Appendix~\ref{app:cross}).

\subsection{Retention of general capabilities}
\label{sec:results_general}

To check for catastrophic forgetting (loss of earlier abilities in further training), we use six benchmarks (MMLU, HellaSwag, ARC-C, Winogrande, TruthfulQA, GSM8K). Their average changes by $-0.071$ from the pre-CPT base to the final SD-DPO model on QuALITY, of which $-0.063$ arises in CPT and only $-0.008$ in SD-DPO (AToKE: $-0.013$). All preference-optimization variants show a similarly small drop in general capability, within measurement error ($\pm 0.5$). Replay (half the training data is general preference data) contributes to this stability, by a share we do not isolate (Appendix~\ref{app:extra}). How the response form changes under each method (the CPT model omits its reasoning, vanilla DPO shortens it further, and SD-DPO states its grounds before answering) is shown in Appendix~\ref{app:length}.

\section{Discussion and Conclusion}
\label{sec:conclusion}

\textbf{What the results show.} On both benchmarks the decisive factor in self-generated data is the share of factually correct rejected responses. Once it dominates, every objective that pushes rejected responses down at initialization (vanilla DPO, IPO, KTO) fails to beat the baseline, while inverting them with a style-cancelling weight succeeds with both scores (cFA $p{\approx}0.33$, pFA $p{=}0.148$). The same contrast appears consistently across the base models we tested, which differ in family and in size (Appendix~\ref{app:cross}).

\textbf{Limitations.} The AToKE results use a single seed, and all judging is by an LLM. With too large a budget SD-DPO repeats to the length cap (2.8\% at the best checkpoint, 14\% at the final; less accurate than the rest); suppressing it without budget control is future work. Appendix~\ref{app:limits} lists all.

\textbf{Conclusion.} Storing knowledge and eliciting it are different optimization problems. A preference stage that respects what actually differs within each pair updates LLM knowledge at a fraction of the CPT cost, and answers with the fact that holds at the queried period.

\subsubsection*{Acknowledgments}
This work was supported by JSPS KAKENHI Grant Number JP24H00727.
Computational resources were provided by the TSUBAME~4.0 supercomputer
at Institute of Science Tokyo (formerly Tokyo Institute of Technology).

\bibliography{references_en}
\bibliographystyle{iclr2027_conference}

\appendix

\section{Theoretical framework}\label{app:theory}

\subsection{Storage and access}\label{app:storage}
We decompose QA accuracy into the product of whether a fact is encoded in the parameters (\emph{storage}) and whether the policy emits it when asked (\emph{access}; what the main text calls elicitation). For fact $j$, let $\mathrm{St}_j(\theta)$ be the event that the fact is stored, and, conditional on it, partition the model's output into the correct answer $\mathrm{El}_j$, a wrong or outdated answer $\mathrm{Stale}_j$, and anything else $\mathrm{Oth}_j$. Then
\begin{equation}\label{eq:decomp}
\Pr[\text{correct}_j] = \underbrace{\Pr[\mathrm{St}_j(\theta)]}_{\text{storage}}\cdot\underbrace{\Pr[\mathrm{El}_j\mid \mathrm{St}_j]}_{\text{access}},\quad
\Pr[\mathrm{El}_j\mid \mathrm{St}_j]=1-\Pr[\mathrm{Stale}_j\mid \mathrm{St}_j]-\Pr[\mathrm{Oth}_j\mid \mathrm{St}_j],
\end{equation}
so accuracy rises whenever $\Pr[\mathrm{Stale}\mid\mathrm{St}]$ falls without $\Pr[\mathrm{Oth}\mid\mathrm{St}]$ rising. CPT maximizes next-token likelihood on the synthetic corpus and thereby increases storage $S(\theta):=\Pr_j[\mathrm{St}_j(\theta)]$ (this is what the EntiGraph scaling law shows), but it does not optimize the probability of emitting the fact when asked (the access factor). In our preference data, about 66\% of the CPT model's responses agree with the gold answer on the facts (cFA $\ge 0.5$, Section~\ref{sec:fa}), showing that CPT has stored many of the facts. Section~\ref{sec:results_quality} shows that standard DPO, which pushes these responses down, does not improve accuracy, whereas SD-DPO, which keeps them and retains only the factual signal, does---consistent with the improvement occurring in the second factor (access).

\subsection{Preference optimization stays near the reference}\label{app:kl}
The RLHF objective $\max_\pi \mathbb{E}[r]-\beta\,\mathrm{KL}(\pi\|\piref)$ has the closed-form maximizer $\pi^\star(y\mid q)\propto\piref(y\mid q)\exp(r(q,y)/\beta)$ \citep{rafailov2023direct}: i.e., the reference policy's probabilities scaled up or down exponentially according to the reward. Because $\pi^\star$ maximizes the objective, evaluating the objective at $\piref$ itself (whose KL term vanishes) gives
$\mathbb{E}_{\pi^\star}[r]-\beta\,\mathrm{KL}(\pi^\star\|\piref)\ge\mathbb{E}_{\piref}[r]$, hence
\begin{equation}\label{eq:klbound}
\mathrm{KL}(\pi^\star\|\piref)\;\le\;\Delta r/\beta,\qquad \Delta r:=\mathbb{E}_{\pi^\star}[r]-\mathbb{E}_{\piref}[r].
\end{equation}
For bounded rewards the right side is $O(1/\beta)$. Assuming that storage $S(\theta)$ depends on $\theta$ only through the output distribution given the question, and is $L_S$-Lipschitz with respect to changes in that distribution measured in total variation (Assumption A1), Pinsker's inequality, which relates KL divergence to total variation, gives
\begin{equation}\label{eq:storage}
\big|S(\theta)-S(\theta_{\text{ref}})\big|\;\le\;L_S\sqrt{\tfrac{1}{2}\mathrm{KL}(\pi_\theta\|\piref)}.
\end{equation}
Thus a preference-optimization stage that stays near $\piref$ changes how knowledge is accessed while approximately preserving what CPT stored. The bound controls the ideal $\pi^\star$; the learned $\pi_\theta$ can overshoot it when trained too long, which is exactly the late-training drop of Section~\ref{sec:budget}. Consistently, the near-zero degradation on general benchmarks (Section~\ref{sec:results_general}) is what \eqref{eq:storage} predicts, though it is corroboration rather than proof.

\subsection{Derivation of Proposition~\ref{prop:cancel}}\label{app:prop}
At $\theta=\theta_{\text{ref}}$ every margin is $\Delta_i=0$, so the sigmoid factor in the gradient of $-\log\sigma(\beta\Delta_i)$ equals $\sigma(-\beta\Delta_i)=1/2$ for all pairs, and the per-pair gradient reduces to $-\tfrac{\beta}{2}\nabla_\theta\Delta_i$ on non-inverted pairs and $+\tfrac{\beta}{2}\nabla_\theta\Delta_i$ on inverted pairs. Substituting $\Delta_i\approx\Delta_i^{(\mathrm{F})}+\Delta^{(\mathrm{S})}$ and summing with the weights of \eqref{eq:sddpo},
\begin{equation}
\nabla_\theta\mathcal{L}_{\text{SD}}\big|_{\theta_{\text{ref}}}\;\propto\;
-\Big[\textstyle\sum_{i:c_i=1}\nabla_\theta\big(\Delta_i^{(\mathrm{F})}+\Delta^{(\mathrm{S})}\big)
-w\sum_{i:c_i=0}\nabla_\theta\big(\Delta_i^{(\mathrm{F})}+\Delta^{(\mathrm{S})}\big)\Big].
\end{equation}
Under Assumption A2 the style term does not depend on the pair, so its total coefficient is $|\{i:c_i{=}1\}|-w\,|\{i:c_i{=}0\}| = Np-\tfrac{p}{1-p}N(1-p)=0$ for any $p\in(0,1)$. Only the factual components remain. The proposition concerns the dataset-averaged gradient at initialization; mini-batch noise is reduced by averaging 32 pairs per step, and drift away from the initial regime is limited by the budget control of Section~\ref{sec:budget}.

\subsection{When Assumption A2 breaks: format-specific style components}\label{app:a2}
Our preference data mixes two answer formats---multiple choice (MCQ) and free text (FT)---whose prompts, response templates, and length distributions differ. The style component then splits by format,
$\Delta^{(\mathrm{S})}=\Delta^{(\mathrm{S}_{\mathrm{MCQ}})}\mathds{1}[\mathrm{MCQ}]+\Delta^{(\mathrm{S}_{\mathrm{FT}})}\mathds{1}[\mathrm{FT}]$,
and a single inverse-ratio weight cancels only the pooled style gradient: if the FT share differs between the $c_i{=}1$ and $c_i{=}0$ groups, a residual per-format style gradient survives. In our standard data the FT share of both groups is close to 50\%, and the group ratios computed on the two formats nearly coincide ($p=0.334$ for MCQ vs.\ $0.336$ for FT); including the within-group normalization of the confidence weights, the residual style gradient at initialization is below $0.5\%$ of one group's style gradient. Training on MCQ only would isolate the format effect and is left to future work.

\subsection{Relation to tie training and spurious-correlation learning}\label{app:ties}
\citet{moya2026spurious} analyze DPO with a log-linear policy whose features split into a causal part $\phi_c$, which concerns content, and a spurious part $\phi_s$, which does not, and show that the optimum over the data distribution assigns non-zero weight to $\phi_s$ through two mechanisms: first, a non-zero mean difference in the spurious features between chosen and rejected, $\mu_s=\mathbb{E}[\Delta\phi_s]$, which is learned from the first gradient step; second, leakage through the correlation between causal and spurious features (the cross-moment $\Sigma_{sc}$). Their remedy, \emph{tie training}, adds pairs with $\Delta\phi_c=0$ and $\Delta\phi_s\neq0$ (ties) labeled uniformly at random; the ties contribute nothing to the mean difference and add curvature (second-order terms of the loss, which act as regularization) only in spurious directions.

SD-DPO maps onto this framework with fact as the causal and style as the spurious feature. Our style-only pairs ($c_i=0$) are ties---same fact, different style---identified automatically by the factual-agreement score rather than constructed by hand, which addresses the open problem of tie construction noted in that work. The two methods treat the ties differently. Random labeling makes the ties' own mean difference vanish, but the spurious mean difference of the non-tie pairs (\emph{strict} pairs in their terminology), $\mu_s^{(P)}$, remains and is only shrunk by the added curvature: the optimum with ties mixed in is $\tilde\theta^\star_{\text{mix}}=\tfrac{2\alpha}{\beta}\Sigma_{\text{mix}}^{-1}\mu^{(P)}$, proportional to $\mu^{(P)}$. SD-DPO instead labels every tie in the inverted direction with weight $w=p/(1-p)$, so that the tie group's mean difference exactly cancels that of the strict group and the total mean difference in style is zero at initialization (Proposition~\ref{prop:cancel}); the inverted pairs also supply curvature in style directions, as ties do. 

Two caveats follow. First, cancelling the mean bias does not remove the leakage through correlation; when fact and style are correlated across pairs, SD-DPO relies on staying in the local regime through budget control rather than on an equilibrium guarantee. Second, \citet{moya2026spurious} note that single-direction labeling injects bias; in SD-DPO the injected bias is deliberately calibrated to cancel the strict pairs' bias, which is why the guarantee depends on Assumption A2, whereas random labeling is robust to A2 violations at the cost of leaving the strict-pair bias in place. A variant that randomizes the labels of style-only pairs, or mixes random and inverted labels, is a natural comparison we have not run.

\section{Implementation details of SD-DPO}\label{app:impl}

\paragraph{Threshold $\thr=0.5$ and STS.} The judge's scoring bands follow the six-level Semantic Textual Similarity (STS) scale \citep{cer2017semeval}: $1.0$ (equivalent, level 5), $0.7$--$0.9$ (differs only in unimportant details, level 4), $0.3$--$0.6$ (partial overlap with important differences, levels 2--3), $0.0$--$0.2$ (not equivalent, levels 0--1). The threshold $0.5$ corresponds to the boundary at which the STS rubric switches from ``equivalent'' (level $\ge 3$) to ``not equivalent'' (level $\le 2$); it separates ``rejected answers the same thing'' from ``rejected answers something else''. Moving the threshold to $0.4$ or $0.6$ reclassifies at most $0.4\%$ of pairs under pFA and $7\%$ under cFA. The $\fmax$ filter uses the same threshold.

\paragraph{The implemented loss.} \Eqref{eq:sddpo} is the ideal form; the implementation combines five additions. (i)~A continuous confidence weight ($1-f_i$ on the factual group, $f_i$ on the inverted group), normalized to mean one within each group so that the cancellation is preserved in expectation. (ii)~The pair of group weights $(1,\,w)$ is rescaled to $(1-p,\,p)$, keeping their ratio while preventing the effective learning rate from drifting relative to the baselines. (iii)~A knowledge-preservation term (a hinge penalty) on inverted pairs keeps the likelihood of the correct rejected response from falling below its value under the reference policy. (iv)~Replayed general-purpose preference data is trained with standard DPO, exempt from inversion and weighting. (v)~The end-of-sequence (EOS) token is excluded from the likelihood sums that define $\Delta_i$. 

Items (i)--(iii) combine into the per-pair loss of \eqref{eq:impl} in the main text. The ratio $p$ and the group means are computed on the synthetic data before replay mixing. Proposition~\ref{prop:cancel} analyzes the ideal form; the normalization in (i) carries the cancellation over to \eqref{eq:impl} in expectation, and the knowledge-preservation term is excluded from the analysis because the argument of its $\max$ is exactly zero (at the kink) at initialization.

\paragraph{Why the EOS token must be excluded.} On inverted pairs, training raises the likelihood of the rejected response and lowers that of the chosen one. Every chosen response ends with the same template (``\texttt{...\ Answer: X.}'' followed by EOS), so with $1-p\approx0.66$ of all pairs inverted, the probability of emitting EOS after an answer is pushed down on most of the data. In preliminary runs the model lost the ability to terminate and repeated the same sentence up to the length limit. Excluding the EOS token from both likelihood sums removes the termination pattern from the preference gradient and eliminates the collapse. 

The exclusion is specific to preference inversion; the compared methods (vanilla DPO, IPO, KTO) never invert and never exhibit the phenomenon, so it is not applied to them. Length-aware objectives---an explicit length margin \citep{park2024disentangling} or equalized token counts \citep{lu2024sampo}---are principled alternatives; we keep the minimal fix so that the comparison isolates the effect of inversion, and report output lengths in Appendix~\ref{app:length}.

\section{Factual-agreement scoring}\label{app:fa}

The judge is \texttt{gpt-5-mini} at temperature 0 with reasoning effort set to low. Judge inputs are English; the prompts below are used verbatim, with placeholders substituted at run time.

\paragraph{pFA prompt (no source document).}
\begin{quote}\scriptsize
\begin{verbatim}
Question: {question}
Reference Answer: {chosen_answer}
Model Response: {rejected}
---
Compare the factual content of the Reference Answer and Model Response.
Ignore differences in format, length, verbosity, or level of detail.
Focus ONLY on whether they provide the same factual answer to the question.
Score 0.0 to 1.0:
- 1.0 = Exactly the same factual answer
- 0.7-0.9 = Same core answer with minor differences in detail
- 0.3-0.6 = Partially overlapping but significant factual differences
- 0.0-0.2 = Completely different or factually wrong answer
Output ONLY the JSON (no explanation before or after):
{"score": <float>, "reason": "<brief one-sentence explanation>"}
\end{verbatim}
\end{quote}

\paragraph{cFA prompt (with source document).}
\begin{quote}\scriptsize
\begin{verbatim}
You are given a source article, a question about that article,
a reference answer ("chosen"), and a model-generated response ("rejected").
--- SOURCE ARTICLE ---
{article_text}
--- END SOURCE ARTICLE ---
Question: {question}
Reference Answer (chosen): {chosen_answer}
Model Response (rejected): {rejected}
---
Evaluate how accurate and appropriate the Model Response is,
**grounded in the source article**.
Consider ALL of the following (not just factual correctness):
- Does the Model Response use the same key terms and phrases as the source?
- Does it correctly capture character motivations, emotions, and relationships?
- Does it reflect the causal relationships and logical flow of events?
- Does it preserve important nuances, qualifications, or hedging?
- Compared to the Reference Answer, does it miss subtle but important details?
Score 0.0 to 1.0: [same bands as above]
Output ONLY the JSON (no explanation before or after):
{"score": <float>, "reason": "<brief one-sentence explanation>"}
\end{verbatim}
\end{quote}
Documents longer than 15{,}000 characters are truncated with a note appended.

\paragraph{The grounding factor of cFA.} The response is split into words and scanned greedily for the longest run (up to 50 words) that matches the lower-cased, whitespace-normalized source verbatim; runs of $\ge 3$ words count toward the matched total. The matched fraction $v_i$ is computed for both responses, and the judge score is multiplied by $\min(1, v_i^l/v_i^w)$: no penalty if the rejected response is at least as grounded in the source wording as the chosen one, an increasing discount as it drifts into generic prose. The factor is needed because the judge prompt instructs the model to ignore differences in detail, which otherwise awards high scores to generic answers that do not draw on the source. It applies to cFA only.

\begin{table}[h]
\centering
\caption{Score distributions of pFA and cFA on the 26{,}291 MCQ pairs (the free-text side behaves similarly). The bottom row is $p=\Pr[f_i<0.5]$, the share of genuinely wrong rejected responses; $1-p$ is the inverted group. The same threshold defines the $\fmax$ filter, so $p$ also equals the post-filter data fraction.}
\label{tab:fadist}
\small
\begin{tabular}{lcc}
\toprule
$f_i$ band & pFA & cFA \\
\midrule
high ($\ge 0.7$) & 89.2\% & 57.6\% \\
mid ($0.3$--$0.7$) & 1.4\% & 17.0\% \\
low ($<0.3$) & 9.4\% & 25.4\% \\
\midrule
$\Pr[f_i<0.5]\;(=p)$ & ${\approx}11\%$ & ${\approx}33\%$ \\
\bottomrule
\end{tabular}
\end{table}

Under pFA about 36\% of the pairs judged ``nearly equivalent'' drop below $0.7$ once the source document is provided: grounding exposes differences in nuance, causality, and character description that pairwise comparison cannot see. This finer judgment---not the resulting group ratio---is why we use cFA on QuALITY (Section~\ref{sec:fa}); AToKE has no long source document, and pFA with $p=0.148$ still supports the cancellation, as Proposition~\ref{prop:cancel} predicts for any $p$.

\section{Data generation}\label{app:data}

\subsection{QuALITY preference data}\label{app:data_quality}
The source material is the EntiGraph synthetic corpus (per-article entity sets, summaries, and pairwise relation texts). Within each article we enumerate entity chains $A$--$B$--$C$: two relation texts sharing a bridge entity $B$ ($A\neq C$, duplicates removed), and sample up to 100 chains per article with seed 42 (26{,}298 chains; 7 generation failures leave 26{,}291 questions). For each chain, \texttt{gpt-5-mini} receives the summary and the two relation texts and writes one 4-way question that \emph{cannot be answered from either text alone}, with (i)~a randomized correct-option position, (ii)~distractors built from entities that occur in the texts but with subtly wrong relations, causality, or attribution, and (iii)~a reasoning trace. 

\paragraph{Two formats.} Each question is instantiated in two formats. The MCQ prompt is identical to the closed-book evaluation prompt (with choices; the system message requires the final line ``\texttt{Answer: X.}'' with X in A--D); the free-text prompt omits the choices and requires the final line to be ``\texttt{Answer:}'' followed by a complete sentence. The chosen response shares the same reasoning in both formats and ends with ``\texttt{Thought process: ...\ Answer: C.}'' (MCQ) or ``\texttt{Thought process: ...\ Answer: <sentence of the correct option>}'' (free text). Rejected responses are sampled from the CPT model under the system message of the respective format (temperature 0.7, top-$p$ 0.9, max 2{,}048 tokens). The result is $26{,}291\times 2 = 52{,}582$ pairs, 21.7M tokens.

\paragraph{Example pair.} As an example, the first pair asks ``How does Charlie emotionally respond to the Chrysler Building's presence in the ruined city?'' (article ``Phone Me in Central Park''; gold C: ``He finds its grandeur hollow and meaningless without living people, deepening his loneliness.''). The chosen response (76 words) is ``Thought process: Charlie is portrayed as the last man, overwhelmed by solitude while surrounded by empty monuments. \ldots\ Answer: C.'' The MCQ rejected response (116 words) is ``Thought process: The narrative describes Charles as the last man on Earth, wandering through a city that is now a graveyard. \ldots\ Answer: C. He finds its grandeur hollow and meaningless without living people, deepening his loneliness.'' with cFA $=1.0$: the rejected response is factually correct and differs from the chosen only in style, so SD-DPO inverts this pair. The free-text rejected response (253 words) is longer and uses its own format, ``[Reading] \ldots\ [Thought process] \ldots\ [Answer] In ``Phone Me in Central Park,'' the Chrysler Building is a silent observer to the desolation experienced by Charles. \ldots'', with cFA $=0.9$, and is inverted as well.

\subsection{AToKE synthetic data: a worked example}\label{app:data_atoke}
Each AToKE edit case is a structured record: subject, relation templates (with and without explicit time), old object with validity period, new object with validity period. For the running example---Brian Kidd, old fact Manchester United F.C.\ (1967--1974), new fact Arsenal F.C.\ (1974--1976)---the four seed sentences are
\begin{quote}\small
new, implicit: ``Brian Kidd's team is Arsenal F.C..''\\
new, explicit: ``From 1974 to 1976, Brian Kidd is a player of Arsenal F.C..''\\
old, implicit: ``Brian Kidd's team \underline{was previously} Manchester United F.C..''\\
old, explicit: ``\underline{From 1967 to 1974}, Brian Kidd \underline{was} a player of Manchester United F.C..''
\end{quote}
Old-fact sentences carry a past marker (``previously'' when time is implicit, past tense when explicit); new-fact sentences are unmarked, matching AToKE's convention that only past-oriented questions are marked. Each seed sentence is paraphrased 10 times without altering entities or time expressions (the multiplier follows \citealp{yamamoto2025nlp,allen2024physics}), and one QA is generated per sentence, giving $44$ QA per case that cover the four evaluated combinations (new/old $\times$ explicit/implicit) in equal measure. Flattened, the CPT corpus over 8{,}819 cases contains 775{,}311 texts (19.7M tokens, 55\% of which are QA); the preference data contains 387{,}287 pairs (9.6M prompt$+$chosen tokens, 14.6M including rejected), scored with pFA ($p=0.148$).

The evaluation asks four questions per case (CES/CRS/HES/HRS; 35{,}276 questions in total), generated from the same templates as the corresponding seed sentences but with independent surface forms. The model answers by free generation, using the chat template without examples (0-shot); \texttt{gpt-5-mini} judges whether the generation contains the expected answer or one of AToKE's aliases. The criterion is: correct if the response includes the expected answer or an alias, even with additional information that does not contradict it; incorrect if it gives a different answer, says it does not know, or contains content that contradicts the expected answer (e.g., a different answer for the queried period), even when the correct answer is also present. 

\paragraph{Responses that mention both periods.} The model often states the fact for the queried period and then facts for other periods as well (e.g., asked about 1997--1998: ``David Brown was a player of Manchester United F.C. from 1997 to 1998. He is a player of Hull City A.F.C. from 2000 to 2002.''). When the other fact is correctly attributed to its own period this is not a contradiction and counts as correct; 6.1\% of the responses judged correct for SD-DPO are of this kind. Conversely, responses that assign the other answer to the queried period yet were judged correct (judge misses) number 11 of 8{,}689 explicit-period questions (0.13\%). 

\paragraph{Comparability and train--test overlap.} Because our rows and the prior-work rows (ROME, MEMIT) use different evaluation protocols, numerical comparisons between them are for reference only. Because questions are template-generated, 1{,}228 of the 35{,}276 evaluation questions (3.48\%) coincide verbatim with a training question; removing them changes overall accuracy by at most 0.001 (CPT 0.884, SD-DPO 0.981), so all reported numbers use the full set.

\section{Hyperparameters and protocols}\label{app:hyper}

\begin{table}[h]
\centering
\caption{Training hyperparameters (QuALITY experiments). $\beta$ applies to DPO and its variants; $\thr$ and $\lambda_{\text{KP}}$ to SD-DPO only. Schedulers: constant without warmup for DPO and its variants, cosine with warmup ratio 0.05 for CPT. Sequence lengths 1024/512 (total/prompt), seed 42, weight decay 0.01, replay from HH-RLHF helpful-base. All training is full-parameter under FSDP on four H100 GPUs. AToKE differs in: CPT lr $2{\times}10^{-5}$, SD-DPO lr $1{\times}10^{-6}$ with cosine scheduler (warmup 0.1).}
\label{tab:hyper}
\small
\begin{tabular}{lccc}
\toprule
 & CPT & Vanilla DPO & SD-DPO \\
\midrule
learning rate & 5e-6 & 1e-6 & 5e-7 (ablations: 1e-6, 2.5e-7) \\
$\beta$ & --- & 0.1 & 0.1 \\
$\thr$ / $\lambda_{\text{KP}}$ & --- & --- & 0.5 / 1.0 \\
$\fmax$ & --- & 0.5 or none & none (ablation: 0.5) \\
replay ratio & 0.1 & 0.5 & 0.5 \\
block size / total batch & 2048 / 16 & --- / 32 & --- / 32 \\
epochs & 2 & 1 & 1 \\
\bottomrule
\end{tabular}
\end{table}

\begin{table}[h]
\centering
\caption{Data and training scale. Step counts are measured from training logs and agree with (examples $\div$ batch size).}
\label{tab:scale}
\small
\begin{tabular}{lrr}
\toprule
 & QuALITY & AToKE \\
\midrule
CPT corpus tokens & 449M & 19.7M \\
CPT steps & 27{,}404 & 1{,}200 \\
preference pairs & 52{,}582 & 387{,}287 \\
preference tokens & 21.7M & 14.6M \\
replay pairs & 43{,}835 & 43{,}835 \\
preference steps & 3{,}014 & 13{,}473 \\
\bottomrule
\end{tabular}
\end{table}

\paragraph{Evaluation protocol.} The QuALITY evaluation set is all 4{,}609 questions about the 265 articles in the public train and dev splits, as in the public EntiGraph implementation (the official test set is not used because its labels are hidden). These articles form the source corpus for CPT, so the closed-book evaluation measures whether their knowledge was stored and can be elicited. 

QuALITY MCQ is evaluated 0-shot with the chat template, $n=8$ samples per question, scoring one uniformly drawn parseable answer (``random valid''). The adopted-answer accuracy equals the ratio of (parseable and correct) to (parseable) probabilities and is therefore independent of $n$. The fraction of parseable generations (\emph{valid-answer rate}) is tracked as an indicator of broken response formatting; the unfiltered vanilla DPO runs drop to 59--84\%, so part of their accuracy loss reflects broken response formatting. 

With a constant learning rate, no warmup, and one epoch, the checkpoint at progress ratio $r$ is equivalent to training on a fraction $r$ of the data: an explicit check at $r{=}0.25$ found a 0.0087 accuracy difference, comparable to rerun-to-rerun noise (${\approx}0.01$).

\paragraph{Baseline selection.} The EntiGraph paper reports 56.22\% (Llama-3-8B base model, 5-shot prompting with chain of thought) but its public implementation, by default, samples $n{=}8$, adopts only the \emph{first} response, and counts only parse-successes in the denominator, which structurally inflates accuracy; under the paper-consistent protocol ($n{=}64$, random valid, all questions in the denominator) our reproduction of CPT starting from the base model drops to about 47\%, and the replay corpus had to be substituted because RedPajama Books is no longer distributed. 

To avoid these instabilities we build on Llama-3-8B-\emph{Instruct} and evaluate everything---including the baseline---with one pipeline (0-shot chat template, $n{=}8$, random valid, closed book). The resulting 56.21\% agrees with the EntiGraph curve prediction at 449M tokens (56.1\%), though this agreement crosses evaluation protocols and is used only for the token-equivalence analysis of Section~\ref{sec:results_quality}.

\section{Additional results}\label{app:extra}

\subsection{AToKE across base models}\label{app:cross}
Table~\ref{tab:atoke_cross_full} applies the data, procedure, and evaluation conditions of the main text to base models of different families and sizes. $p$ is the share of pairs whose rejected response is factually wrong, measured per model from that model's own post-CPT responses. On all three models tested here, SD-DPO exceeds both CPT only and vanilla DPO, and vanilla DPO falls below CPT only. The size of that drop tracks $p$: it is largest where factually correct rejected responses are most common (Qwen3, $p=0.132$: $-0.359$; Gemma4, $p=0.734$: $-0.011$). The gain of SD-DPO instead tracks the CPT-only level, and is largest where storage is weakest and headroom largest (Gemma4 $+0.232$, Llama $+0.097$, Qwen3 $+0.046$). Inversion also works at large $p$: on Gemma4 only a quarter of the pairs are inverted, yet the weight $w=p/(1-p)$ balances the two groups, so the style gradients still cancel.

Among the individual metrics, only CES on Qwen3 falls below CPT only (0.901 to 0.870). It is an exchange: 1{,}001 questions that CPT answered correctly are lost and 725 that it missed are gained. Of the lost responses, 71.8\,\% name \emph{the answer of a different edit case} and only 5.0\,\% name the old fact of the same case; the model does not confuse the two periods, it replaces a stored fact with another case's answer. CES is Qwen3's highest metric of the four (0.901), so the headroom is small and this exchange comes out negative, whereas on the other three the same exchange is positive (CRS $+781$, HES $+603$, HRS $+512$ questions) and Overall rises.

\begin{table}[h]
\centering
\caption{The single-edit setting of AToKE across base models (accuracy 0--1). The Llama rows are those of Table~\ref{tab:atoke}. ``Len.'' is the mean response length in words at evaluation.}
\label{tab:atoke_cross_full}
\footnotesize
\setlength{\tabcolsep}{3.2pt}
\begin{tabular}{llcccccccccc}
\toprule
Base model & $p$ & Method & CES & HES & CRS & HRS & Current & History & Overall & Len. \\
\midrule
Llama-3.1-8B & 0.148 & CPT only & 0.905 & 0.782 & 0.880 & 0.974 & 0.892 & 0.878 & 0.885 & 6 \\
 & & $+$ Vanilla DPO & 0.608 & 0.823 & 0.633 & 0.887 & 0.620 & 0.855 & 0.738 & 40 \\
 & & $+$ SD-DPO & 0.985 & 0.954 & 0.989 & 0.998 & 0.987 & 0.976 & \textbf{0.982} & 272 \\
\midrule
Qwen3-8B & 0.132 & CPT only & 0.901 & 0.782 & 0.804 & 0.890 & 0.853 & 0.836 & 0.844 & 7 \\
 & & $+$ Vanilla DPO & 0.368 & 0.530 & 0.356 & 0.685 & 0.362 & 0.608 & 0.485 & 3 \\
 & & $+$ SD-DPO & 0.870 & 0.850 & 0.893 & 0.948 & 0.881 & 0.899 & \textbf{0.890} & 33 \\
\midrule
Gemma4-E2B & 0.734 & CPT only & 0.691 & 0.590 & 0.678 & 0.775 & 0.685 & 0.683 & 0.684 & 10 \\
 & & $+$ Vanilla DPO & 0.625 & 0.635 & 0.633 & 0.796 & 0.629 & 0.716 & 0.673 & 3 \\
 & & $+$ SD-DPO & 0.923 & 0.870 & 0.909 & 0.961 & 0.916 & 0.916 & \textbf{0.916} & 10 \\
\bottomrule
\end{tabular}
\end{table}

\begin{table}[h]
\centering
\caption{Seed robustness of the main configuration (three training seeds; data, initialization, and hyperparameters fixed). QuALITY MCQ \%.}
\label{tab:seed}
\small
\begin{tabular}{lcccc}
\toprule
 & seed 42 & seed 43 & seed 44 & mean $\pm$ SD \\
\midrule
SD-DPO best (50\% usage) & 59.17 & 59.91 & 56.57 & 58.55 $\pm$ 1.75 \\
SD-DPO final (100\%) & 57.98 & 55.79 & 56.73 & 56.83 $\pm$ 1.10 \\
\bottomrule
\end{tabular}
\end{table}

\begin{table}[h]
\centering
\caption{All combinations for vanilla DPO (best\,/\,final MCQ \%, seed 42). No cell clearly exceeds the 56.21\% baseline; the best, pFA $\times$ $2.5{\times}10^{-7}$, is $+0.58$, inside the standard deviation across seeds (1.75, Table~\ref{tab:seed}). Valid-answer rates in all filtered cells exceed 98.5\%.}
\label{tab:matrix}
\small
\begin{tabular}{lccc}
\toprule
$\fmax$ & lr $10^{-6}$ & lr $5{\times}10^{-7}$ & lr $2.5{\times}10^{-7}$ \\
\midrule
none & 52.77\,/\,47.01 & 56.01\,/\,54.64 & 56.53\,/\,56.27 \\
pFA (11.7\% kept) & 56.45\,/\,55.99 & 56.55\,/\,56.30 & \textbf{56.79\,/\,56.56} \\
cFA (33.6\% kept) & 55.29\,/\,48.84 & 56.13\,/\,55.64 & 56.19\,/\,56.15 \\
\bottomrule
\end{tabular}
\end{table}

\begin{table}[h]
\centering
\caption{General capabilities. Left: QuALITY experiments (Llama-3-8B-Instruct; MMLU/Winogrande/TruthfulQA \texttt{acc}, HellaSwag/ARC-C \texttt{acc\_norm}, GSM8K strict-match). Right: AToKE experiments (Llama-3.1-8B-Instruct; GSM8K flexible-extract, strict-match in text). ``vs Base'' is the total change; ``vs CPT'' isolates the SD-DPO stage.}
\label{tab:gen}
\small
\setlength{\tabcolsep}{3.6pt}
\begin{tabular}{lccccc@{\hskip 14pt}ccccc}
\toprule
 & \multicolumn{5}{c}{QuALITY experiments} & \multicolumn{5}{c}{AToKE experiments} \\
\cmidrule(lr){2-6}\cmidrule(lr){7-11}
Task & Base & CPT & SD-DPO & vs Base & vs CPT & Base & CPT & SD-DPO & vs Base & vs CPT \\
\midrule
MMLU & 0.647 & 0.614 & 0.612 & $-0.035$ & $-0.002$ & 0.683 & 0.638 & 0.629 & $-0.054$ & $-0.009$ \\
HellaSwag & 0.757 & 0.764 & 0.757 & $\pm0.000$ & $-0.007$ & 0.795 & 0.755 & 0.750 & $-0.045$ & $-0.005$ \\
ARC-C & 0.554 & 0.497 & 0.507 & $-0.047$ & $+0.010$ & 0.556 & 0.556 & 0.567 & $+0.011$ & $+0.010$ \\
Winogrande & 0.711 & 0.721 & 0.718 & $+0.007$ & $-0.002$ & 0.736 & 0.732 & 0.725 & $-0.011$ & $-0.007$ \\
TruthfulQA & 0.525 & 0.444 & 0.418 & $-0.107$ & $-0.026$ & 0.546 & 0.459 & 0.438 & $-0.108$ & $-0.021$ \\
GSM8K & 0.756 & 0.531 & 0.511 & $-0.245$ & $-0.020$ & 0.773 & 0.518 & 0.474 & $-0.299$ & $-0.044$ \\
\midrule
mean & 0.658 & 0.595 & 0.587 & $-0.071$ & $-0.008$ & 0.682 & 0.610 & 0.597 & $-0.085$ & $-0.013$ \\
\bottomrule
\end{tabular}
\end{table}

\paragraph{Length asymmetry and output length.}\label{app:length}
Table~\ref{tab:length} summarizes response lengths in the preference data and in model outputs. Two facts matter for interpreting the results. First, chosen and rejected responses differ greatly in length: on QuALITY the rejected response is longer than the chosen one in 98--99.5\% of pairs, by a factor of 2.2--3.5 on average. This is the regime in which \citet{lu2024sampo} show that standard DPO underestimates the chosen reward and can collapse for reasons unrelated to content; it is why we compare against SamPO. This length difference does \emph{not}, however, separate the two groups that SD-DPO treats differently: rejected responses in the factual-difference group and in the style-only group have nearly the same length on AToKE as well, whatever the base model (Llama 8 vs 7 words, Qwen3 10 vs 11, Gemma4 15 vs 13), and on QuALITY (146 vs.\ 134 words on MCQ), so the $c_i$ split is not a length proxy and the ablation that removes only the inversion (Section~\ref{sec:results_quality}(c)) cannot be explained by length. 

\paragraph{Response differences across methods.} In the MCQ format, chosen and rejected share the ``Thought process: \ldots\ Answer: X.'' format, and the evaluation prompt asks for reasoning first. The CPT model skips reasoning and emits only ``Answer: X.'' (28 words); vanilla DPO goes further (15 words), since pushing down long, correct rejected responses also pushes down stating reasons. SD-DPO moves the other way: it recalls the article, states grounds, and answers (200 words at the best point); on the 443 questions only SD-DPO gets right, the grounds typically contain the answer. With too large a budget this tendency overshoots: the model states the answer, then repeats a sentence to the length cap (Section~\ref{sec:conclusion}).

\paragraph{Output length.} Second, SD-DPO makes the model markedly more verbose. Vanilla DPO moves outputs toward the short chosen template (15 words), IPO, KTO and SamPO even more so (2 words, i.e., the bare answer), while SD-DPO moves them toward the long free-form rejected responses (200 words at the best checkpoint, 382 at the final one), with 3--14\% of generations exceeding 1{,}000 words. 

\paragraph{Generations that hit the length cap.} On AToKE extremely long generations are more frequent (18\% above 1{,}000 words) but these answers are still judged correct at 0.988, since the judge checks containment of the expected answer. On QuALITY the opposite holds: generations that hit the cap at the final checkpoint (14.0\% of all) keep the answer format less often (79.8\% vs.\ 95.6\% for normal generations) and, when they do, are correct less often (44.5\% vs.\ 60.0\%). Inspecting them, the ``Answer: X'' line appears near the start (median at 6\% of the length): the model states the answer, fails to stop, and repeats an explanatory sentence. For example, one final-checkpoint response (gold B, 1{,}741 words) answers ``Thought process: \ldots\ Answer: B. The article suggests \ldots'' and then repeats the sentence ``The article does not provide a direct answer to the question \ldots, but it does provide context \ldots'' 23 times. At the best checkpoint (usage 50\%) such generations are 2.8\%. This is the out-of-distribution length extrapolation described by \citet{park2024disentangling}, here in the direction selected by inversion; excluding the EOS token reduces such generations relative to our preliminary runs, in which most outputs hit the cap, but does not eliminate them. Inspecting the generations that run to the length cap shows their cause: on AToKE, of 300 answers longer than 1{,}000 words, 95\% of sentences are repeats of one or two sentences (e.g., ``He was a player of Hungary national football team from 1996 to 1997.'' repeated 113 times), so the extremely long generations are the repetition collapse of Appendix~\ref{app:impl} surviving in attenuated form, not additional content.

\paragraph{A worked example.} Table~\ref{tab:example} shows the first of eight sampled answers from each model on one QuALITY question that only SD-DPO answers correctly (all eight samples agree within each model). The CPT model and vanilla DPO answer without reasoning and pick a wrong option; SD-DPO recalls the story, and the correct answer appears inside the recalled grounds. Over the test set, 443 questions are answered correctly ($\geq 7/8$) by SD-DPO and incorrectly ($\leq 1/8$) by both baselines, against 303 in the opposite direction; when SD-DPO is wrong it typically narrates a coherent but incorrect version of the story with the same confidence, so longer reasoning is a route to the stored knowledge, not a guarantee of reaching it.

\begin{table}[h]
\centering
\caption{Responses to one QuALITY question (\emph{Break a Leg}, Jim Harmon, 1958; labeled ``hard''). \textbf{Q:} Why couldn't Baxter use his own bathroom? \textbf{A.} It wasn't nice enough \quad \textbf{B.} It was out of order \quad \textbf{C.} He thought it was too nice for him \quad \textbf{D.} He was trying to sneak off the ship (correct). Each row is the first of eight samples; the bracketed letters are the options chosen by all eight.}
\label{tab:example}
\small
\begin{tabular}{@{}p{0.16\linewidth}p{0.80\linewidth}@{}}
\toprule
CPT only [B$\times$8] & Answer: B. It was out of order \\
\midrule
Vanilla DPO [B$\times$8] & Answer: B. \\
\midrule
SD-DPO [D$\times$8] & [Thought process] In the story ``Break a Leg'' by Jim Harmon, the narrative revolves around the character Charlie Baxter, an Accident Prone individual who is part of a spaceship crew. The spaceship is equipped with a luxurious Admiralty-style stateroom, which includes a private bathroom. However, Baxter is not allowed to use his own bathroom. This is because the door to his bathroom is locked, and the only way to access it is through the captain's bathroom. The captain has the key to Baxter's bathroom, and he only gives it to Baxter when he is summoned. This arrangement is a security measure to prevent Baxter from attempting to leave the ship without permission, as he is prone to accidents and could potentially harm himself or others. [Answer] Answer: D. He was trying to sneak off the ship. \\
\bottomrule
\end{tabular}
\end{table}

\begin{table}[h]
\centering
\caption{Response lengths (words). Top: preference pairs; ``rej.\ longer'' is the share of pairs whose rejected response is longer than the chosen one, and the last two columns give the mean rejected length in the factual-difference ($c_i{=}1$) and style-only ($c_i{=}0$) groups. The three AToKE rows are split by the model that generated the rejected responses. Bottom: model outputs on the evaluation sets (QuALITY: $8\times4{,}609$ generations; AToKE: 35{,}276 answers; for base models other than Llama see the ``Len.'' column of Table~\ref{tab:atoke_cross_full}); ``valid'' is the parseable-answer rate and ``$>$1k'' the share of generations exceeding 1{,}000 words.}
\label{tab:length}
\small
\setlength{\tabcolsep}{4pt}
\begin{tabular}{lrrrrr}
\toprule
\multicolumn{6}{l}{\emph{Preference pairs}} \\
 & chosen & rejected & rej.\ longer & rej.\ ($c_i{=}1$) & rej.\ ($c_i{=}0$) \\
\midrule
QuALITY MCQ (cFA) & 61 & 135 & 98.1\% & 146 & 134 \\
QuALITY free text (cFA) & 73 & 254 & 99.5\% & 269 & 251 \\
AToKE (pFA, Llama) & 3 & 7 & 57.3\% & 8 & 7 \\
AToKE (pFA, Qwen3) & 3 & 11 & 92.4\% & 10 & 11 \\
AToKE (pFA, Gemma4) & 3 & 14 & 100.0\% & 15 & 13 \\
\midrule
\multicolumn{6}{l}{\emph{Model outputs}} \\
 & mean & median & valid & $>$1k & MCQ / Overall \\
\midrule
QuALITY: CPT only & 28 & 11 & 95.4\% & 0.1\% & 56.21 \\
QuALITY: vanilla DPO ($5{\times}10^{-7}$, final) & 15 & 5 & 97.1\% & 0.4\% & 54.64 \\
QuALITY: vanilla DPO ($10^{-6}$, final) & 223 & 35 & 63.9\% & --- & 47.01 \\
QuALITY: IPO (final) & 19 & 7 & 54.8\% & --- & 45.98 \\
QuALITY: KTO (final) & 2 & 2 & 100.0\% & --- & 54.47 \\
QuALITY: SamPO (final) & 2 & 2 & 99.6\% & 0.0\% & 57.32 \\
QuALITY: SD-DPO (best, 50\%) & 200 & 140 & 93.1\% & 2.8\% & 59.17 \\
QuALITY: SD-DPO (final) & 382 & 150 & 93.4\% & 14.0\% & 57.98 \\
AToKE: CPT only & 6 & 4 & --- & 0.0\% & 0.885 \\
AToKE: vanilla DPO & 40 & 3 & --- & 1.9\% & 0.738 \\
AToKE: SD-DPO & 272 & 10 & --- & 18.4\% & 0.982 \\
\bottomrule
\end{tabular}
\end{table}

\paragraph{Capability trajectories.} Measuring the average over the six benchmarks (AVG6) at every intermediate checkpoint shows that all runs trained on unfiltered data to 21.7M tokens end in the same narrow range ($-1.36$ to $-1.63$ points below the 55.99\% baseline, standard error ${\approx}\pm0.5$ per point), regardless of learning rate or loss---while their QuALITY accuracies differ by more than 10 points (57.98 vs.\ 50.56 vs.\ 47.01). The only run with a small AVG6 drop ($-0.21$), vanilla at $5{\times}10^{-7}$, is also the run that barely moves on QuALITY. 

Capability loss therefore tracks how far the model actually moved, not how many tokens were consumed, consistent with \eqref{eq:storage}; and the task-accuracy drop is a separate, larger-scale phenomenon that the learning rate alone removes. The dominant contributor throughout is GSM8K; HellaSwag and Winogrande stay within $\pm0.02$ in every condition.

\paragraph{IPO, KTO and SamPO details.} Under identical conditions (same data, cFA, no filter, replay 0.5, learning rate $5{\times}10^{-7}$, seed 42), IPO reaches best/final $55.68/45.98$, KTO $56.05/54.47$ and SamPO $57.32/57.32$, versus vanilla $56.01/54.64$ and SD-DPO $\mathbf{59.17/57.98}$. SamPO is the only baseline that exceeds the CPT baseline; its four checkpoints lie in a narrow range ($56.54$, $57.00$, $56.89$, $57.32$ at $10/25/50/100\%$ of the data), with the full pass best, and it shows no late drop---consistent with its reward being length-normalized. 

\paragraph{How SamPO equalizes length.} The equalization works as follows. The DPO reward is the sum over a response of per-token log-likelihood ratios against the reference policy; the number of summands equals the token count, so longer responses obtain larger sums. SamPO leaves the input and the forward pass on the complete responses unchanged and equalizes only the number of summands used for the reward: for each pair it takes the shorter length $T_m=\min(T_w,T_l)$ of the chosen ($T_w$) and rejected ($T_l$) responses, draws $T_m$ token positions uniformly at random (without replacement) from each response, sums the log-likelihoods over those positions only, and computes the loss with the usual DPO formula. All positions of the shorter response are used and some summands of the longer one are dropped, but the response text itself is never truncated. The positions are re-drawn at every step, and the policy and reference forward passes on the same batch use identical positions. Down-sampling rather than dividing by the token count follows the original paper, which argues that averaging shrinks the reward scale to $1/T$ and thereby changes the meaning of $\beta$ (its corrected variant DPO-SANorm multiplies back by $(T_w+T_l)/2$) and removes the variance across tokens, and reports that sampling performs slightly better. Implementation and verification: vanilla DPO, IPO, and SD-DPO run in one trainer built on TRL's \texttt{DPOTrainer}, whose forward pass and log-likelihoods are TRL's; vanilla DPO is the degenerate case of our loss and coincides with TRL's sigmoid loss, and IPO applies TRL's formula to TRL's length-normalized log-likelihoods. KTO uses TRL's \texttt{KTOTrainer} with its loss unchanged. SamPO has no TRL implementation, so we implemented the down-sampling as an option of the same trainer and verified it in two ways: with the option off, the trainer reproduces TRL's log-likelihoods exactly (checked during training, $\max|\Delta|=0$ over eight checks), and a unit test of the selection mask confirms that each side receives exactly $T_m$ positions inside the valid range, that the shorter response is used in full, that the policy and reference passes on the same batch use identical positions, and that positions are re-drawn at every step (\texttt{tests/test\_sampo\_selection.py} in the released code). 

\paragraph{Why SamPO's outputs shrink to two words.} We interpret the shrinking of SamPO's outputs to the two-word ``\texttt{Answer: X.}'' as follows: after down-sampling, the long rejected response is no longer pushed down in proportion to its length, leaving only the gradient toward the chosen side, so stating reasons brings no benefit and the model converges to the shortest response that satisfies the answer format (the causal link is untested). 

\paragraph{Choice of IPO and KTO, and excluded variants.} IPO was chosen as the loss-shape representative: it regresses the preference margin $\Delta_i$ toward a finite target $1/(2\beta)$ with the squared error $(\Delta_i-\tfrac{1}{2\beta})^2$, the standard remedy for DPO's weak regularization; near initialization it updates in the same direction as DPO, and the direction reverses for pairs whose margin exceeds the target. KTO represents the data-structure axis, abandoning pair comparison altogether. ORPO and SimPO do without the reference policy, which in our setting is the anchor that preserves stored knowledge (Eq.~\ref{eq:storage}), so their loss-shape effect cannot be isolated; rDPO assumes label noise, which our construction rules out. Each baseline was run at the single shared learning rate; per-method tuning is future work.

\paragraph{General capabilities.} Table~\ref{tab:gen_cross} reports the six benchmarks of Table~\ref{tab:gen} under the same measurement conditions. The drop added by the preference stage is $+0.002$ on Qwen3 and $-0.021$ on Gemma4, comparable to the $-0.013$ of the Llama AToKE run: raising AToKE accuracy substantially while leaving general capability nearly intact does not depend on the base model. We omit the pre-CPT base row. Under this protocol---no chat template, scoring the log-likelihood of each option---the untrained Gemma4-E2B-it scores at chance on MMLU (0.290), below its own post-CPT value (0.528). Transformers and vLLM give the same value (0.2899 and 0.2894), so this is not an implementation problem but the effect of scoring a chat-tuned model as a plain continuation; CPT on plain text restores ordinary language-model behaviour, which makes the pre/post comparison meaningless. Since the quantity of interest is the change added by the preference stage, we compare against CPT only.

\begin{table}[h]
\centering
\caption{General language capability across base models (the six benchmarks and measurement conditions of Table~\ref{tab:gen}; \texttt{acc} for MMLU/Winogrande/TruthfulQA, \texttt{acc\_norm} for HellaSwag/ARC-C, flexible-extract for GSM8K). ``vs CPT'' is the difference from CPT only.}
\label{tab:gen_cross}
\small
\setlength{\tabcolsep}{4pt}
\begin{tabular}{lccccccrr}
\toprule
Model / method & MMLU & HellaS. & ARC-C & Winog. & TQA & GSM8K & Avg. & vs CPT \\
\midrule
Qwen3-8B \quad CPT only & 0.719 & 0.758 & 0.613 & 0.730 & 0.413 & 0.806 & 0.673 & --- \\
\quad $+$ Vanilla DPO & 0.728 & 0.751 & 0.637 & 0.724 & 0.436 & 0.816 & 0.682 & $+0.009$ \\
\quad $+$ SD-DPO & 0.724 & 0.755 & 0.634 & 0.732 & 0.393 & 0.809 & 0.675 & $+0.002$ \\
\midrule
Gemma4-E2B \quad CPT only & 0.528 & 0.639 & 0.497 & 0.627 & 0.429 & 0.495 & 0.536 & --- \\
\quad $+$ Vanilla DPO & 0.532 & 0.635 & 0.525 & 0.626 & 0.420 & 0.495 & 0.539 & $+0.003$ \\
\quad $+$ SD-DPO & 0.498 & 0.632 & 0.490 & 0.620 & 0.404 & 0.449 & 0.515 & $-0.021$ \\
\bottomrule
\end{tabular}
\end{table}

\section{Limitations}\label{app:limits}

\begin{itemize}
\setlength{\itemsep}{1pt}
\item The base models are three families on AToKE (Llama-3.1 8B, Qwen3 8B, Gemma4 E2B; Appendix~\ref{app:cross}) and one on QuALITY (Llama-3 8B Instruct). QuALITY uses one family by design: the purpose there is a comparison on EntiGraph's synthetic data and benchmark, and EntiGraph itself was evaluated on a single model. The starting model and evaluation procedure differ from EntiGraph's as described above, so the comparison is between two models of our own. All three families are transformer-based models of 8B parameters or fewer; larger models and other architectures are untested. Robustness to other model families is a matter for the AToKE side, our main goal.
\item Two benchmarks (QuALITY, AToKE); domain and format diversity is limited.
\item True iterative DPO---re-sampling rejected responses from the current policy each round---is untested; we only tried sequential training on four data splits with a different loss.
\item Hyperparameter search is local, centered on defaults.
\item The EOS exclusion is applied only to SD-DPO, where inversion makes it necessary; its effect on the baselines is unmeasured.
\item Judge scores and AToKE correctness are LLM-judged, without human verification.
\item The AToKE experiments use a single random seed (42).
\item Numerical comparisons with prior work on AToKE (ROME, MEMIT, METO) are for reference only: the base model and the evaluation protocol differ (the original paper does not publish how its accuracies are computed; we judge free generations with an LLM).
\item 3.48\% of AToKE evaluation questions coincide verbatim with training questions; removing them changes accuracy by $\le 0.001$, but the split is not fully disjoint.
\item Noise-robust DPO variants (rDPO, $\beta$-DPO) are compared conceptually, not empirically, because their label-noise premise does not hold in our construction.
\item Data-side regularization (e.g., generative replay) as an alternative to the KL constraint toward the reference policy is not compared.
\item General-capability retention partly reflects the 0.5 replay ratio; the contributions of replay and of the KL constraint are not separated.
\item At large training budgets SD-DPO becomes verbose: at the final checkpoint 14\% of generations repeat to the length cap and are less accurate than normal generations (Appendix~\ref{app:length}). Suppressing this with length-aware objectives or stopping control is untested.
\item The pFA variant of SD-DPO on QuALITY is unmeasured; the score choice is justified by distribution and judging content only.
\item The CPT-vs-elicitation efficiency comparison rests on one point (449M tokens) of the scaling curve.
\item Proposition~\ref{prop:cancel} removes the \emph{mean} of the style gradient. The second mechanism of \citet{moya2026spurious}---leakage through correlation between factual and style features---is not addressed; SD-DPO relies on staying near initialization rather than on an equilibrium guarantee (Appendix~\ref{app:ties}).
\item Assumption A2 is also violated by variation in rejected length: the style component, a sum of per-token log-likelihoods, scales with the number of tokens \citep{lu2024sampo}, and rejected responses vary widely in length (Appendix~\ref{app:length}).
\item SD-DPO lengthens outputs substantially and leaves some extremely long, run-away generations (Appendix~\ref{app:length}); combining inversion with explicit length control is untested.
\item The length-debiased comparison uses SamPO at the single shared learning rate; other length-aware objectives are not compared.
\item Improving CPT itself (better synthetic data generation) is out of scope.
\end{itemize}

\end{document}